\documentclass[preprint,3p,times,twocolumn]{elsarticle}           

\usepackage{amssymb}

\def\appendixname{}        

\usepackage{microtype}
\usepackage{graphicx}
\usepackage{subfigure}
\usepackage{booktabs} 
\usepackage{kotex} 

\usepackage{hyperref}

\usepackage{amsmath}

\usepackage{xcolor}

\usepackage[capitalize,noabbrev]{cleveref}
\usepackage{cases}

\usepackage{booktabs}

\usepackage{algorithm}  
\usepackage{algorithmic}

\journal{Image and Vision Computing}

\begin{document}
\begin{frontmatter}


\title{One-shot Ultra-high-Resolution Generative Adversarial Network \\ That Synthesizes 16K Images On A Single GPU}

\author[1]{Junseok Oh}
\ead{junseokoh96@gmail.com}

\author[1]{Donghwee Yoon}
\ead{yoondonghwee@gmail.com} 



\author[1]{Injung Kim\corref{cor1}}
\ead{ijkim@handong.edu}

\cortext[cor1]{Corresponding author}

\affiliation[1]{organization={School of CSEE, Handong Global university},
            city={Pohang},
            postcode={37554}, 
            country={Republic of Korea}}

\begin{abstract}
We propose a one-shot ultra-high-resolution generative adversarial network (OUR-GAN) framework
that generates non-repetitive 16K ($16,384\times8,640$) images from a single training image and is trainable on a single consumer GPU. OUR-GAN generates an initial image that is visually plausible and varied in shape at low resolution, and then gradually increases the resolution by adding detail through super-resolution. Since OUR-GAN learns from a real ultra-high-resolution (UHR) image, it can synthesize large shapes with fine details and long-range coherence, which is difficult to achieve with conventional generative models that rely on the patch distribution learned from relatively small images. OUR-GAN can synthesize high-quality 16K images with 12.5 GB of GPU memory and 4K images with only 4.29 GB as it synthesizes a UHR image part by part through seamless subregion-wise super-resolution. Additionally, OUR-GAN improves visual coherence while maintaining diversity by applying vertical positional convolution. In experiments on the ST4K and RAISE datasets, OUR-GAN exhibited improved fidelity, visual coherency, and diversity compared with the baseline one-shot synthesis models. To the best of our knowledge, OUR-GAN is the first one-shot image synthesizer that generates non-repetitive UHR images on a single consumer GPU. The synthesized image samples are presented at https://our-gan.github.io.
\end{abstract}




\end{frontmatter}

\begin{figure*}[ht]
\vskip 0.1in
\begin{center}
\centerline{\includegraphics[width=0.95\textwidth]{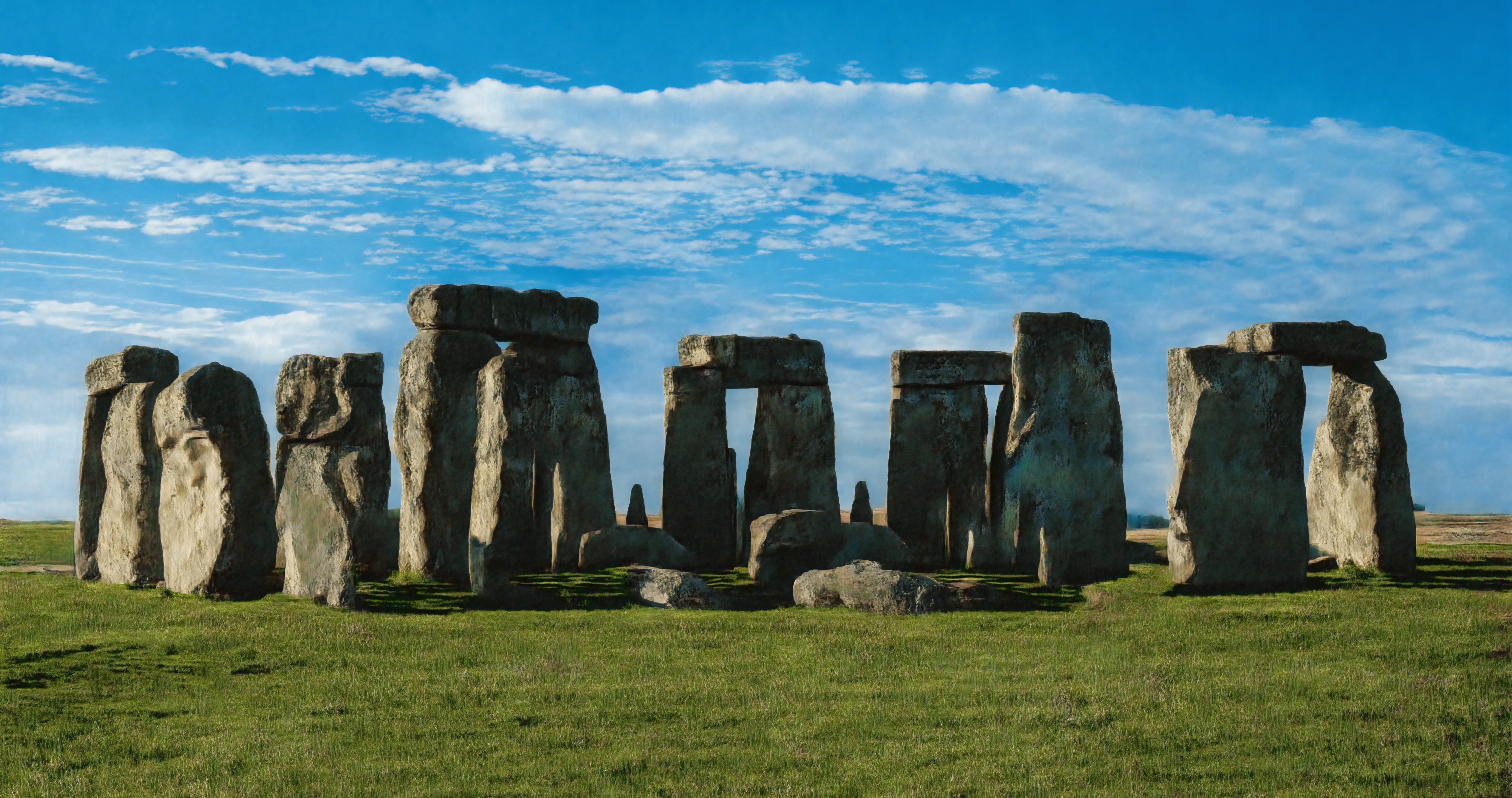}}
\caption{A 16K ($16,384\times8,640$) UHR image synthesized by OUR-GAN on a single Titan RTX GPU. The model was trained with a single 8K image. OUR-GAN can synthesize diverse UHR images with high fidelity and visual coherence from a single training image whose resolution is $2-4\times$ smaller than the target resolution.}
\label{fig:16K_Image_Synthesis}
\end{center}
\vskip -0.4in
\end{figure*}

\section{Introduction}
\label{Introduction}
Recently developed generative adversarial network (GAN) models synthesize high-quality images by applying a variety of techniques such as the progressive growing approach \cite{PGGAN, StyleGANv1, StyleGANv2}, a hierarchy of multi-scale generators and discriminators \cite{StackGAN, StackGAN++, HDGAN, LAPGAN, SinGAN, ConSinGAN, HP-VAE-GAN}, and auxiliary losses \cite{PGGAN, StackGAN++, HDGAN, StyleGANv1, StyleGANv2}.
However, most of the existing generative models have limited output resolutions and require a large volume of training data.
In contrast, there is an increasing demand for synthesizing ultra-high-resolution (UHR) images with 4K or higher resolutions. 
However, collecting a large number of UHR images is costly and even impossible in certain application fields, such as micrographics and material patterns.
For example, while a variety of marble patterns are required for use as interior building materials, acquiring a large volume of high-resolution marble images requires significant expense and effort.

One-shot image synthesis is the task of generating diverse images from only a single training example, and the synthesized images should have similar properties to the training example. It is a sub-field of one-shot learning that aims to train a model with only a single example, and is particularly useful in scenarios where obtaining large amounts of labeled data is costly or impractical.
A one-shot image synthesizer should generate images with contents and styles similar to the training image but with different shapes, and the synthesized images should be visually coherent and plausible. In addition, since synthesizing UHR images requires a large amount of computation and memory, one-shot UHR image synthesizers should be efficient to operate in general computational environments. It is challenging to satisfy all those requirements.
Most existing one-shot synthesis models generate diverse images with similar properties to the training image based on the internal patch distribution of the training image \cite{SinGAN, ConSinGAN, InGAN, HP-VAE-GAN, GPNN, GPDM, ExSinGAN, PetsGAN, SinFusion, SinDiffusion, SinDDM}. However, these models have limited output resolution in practice and their visual coherence and diversity have room for improvement.

Compared with a 1K ($1,024\times1,024$) image, a 4K ($4,096\times2,160$) image contains $8\times$ more pixels and a 16K ($16,384\times8,640$) image contains $128\times$ more pixels. Therefore, the UHR image synthesizer should produce substantially more information than ordinary image synthesizers. To generate a UHR image with limited GPU memory, the model should synthesize images part by part and then concatenate them into a full-size image. However, it is challenging to maintain fidelity, visual coherence, and shape diversity while synthesizing UHR images by subregion.

Few studies have explored non-repetitive UHR image synthesis with a limited amount of GPU memory \cite{ALIS, InfinityGAN, TamingTransformer}.
They synthesize images of arbitrary size by concatenating patterns learned from ordinary-sized images. However, it is hard to synthesize shapes larger than those of the training images with such models. 
Moreover, they require a large volume of training data.

In this study, we propose a one-shot ultra-high-resolution generative adversarial network (OUR-GAN), that synthesizes non-repetitive high-fidelity UHR images, as shown in \cref{fig:16K_Image_Synthesis}, and is trainable with a single training image on a single consumer GPU. Unlike existing models that generate UHR images by connecting small patches, OUR-GAN synthesizes diverse and visually coherent images at low resolution and then gradually increases the resolution by adding detail through super-resolution. OUR-GAN synthesizes UHR images with limited GPU memory while preventing discontinuity at subregion boundaries by seamless subregion-wise super-resolution. 
More importantly, OUR-GAN can synthesize not only large-sized images but also images with visually coherent large shapes and fine detail, as presented in \cref{fig:16K_Image_Synthesis}, which is not easily achieved with conventional patch-based methods.

Additionally, we improved visual coherence while maintaining diversity by exploiting the spatial bias of scenery images through vertical coordinate convolution. Up to our knowledge, OUR-GAN is the first non-repetitive UHR image synthesis model that is trainable with a single image on a single GPU (first appeared in arXiv 2022 \cite{OUR-GAN_arXiv}).
A few datasets \cite{RAISE, UHDSR4K, DIV8K} contain UHR images with 4K or higher resolutions. However, as they were not collected for UHR image synthesis, they contains many low-quality or low-resolution images.
For a more reliable evaluation, we introduce a new dataset \textbf{S}cenery and \textbf{T}exture-\textbf{4K} (\textbf{ST4K}) for one-shot UHR image synthesis. 
ST4K consists of 25 images of diverse natural and urban scenery, as well as 25 images of high-quality textures.
Compared with conventional methods, OUR-GAN exhibits improved experimental results on ST4K and RAISE datasets in terms of fidelity, visual consistency, and diversity.
The main contributions of our study are as follows: 
\begin{enumerate}
    \item We propose the first framework to synthesize non-repetitive high-fidelity UHR images and is trainable with a single image on a single GPU.
    \item We present a seamless subregion-wise super-resolution method that prevents discontinuity at subregion boundaries with minimal increase in computational and memory requirements.
    \item We apply vertical coordinate convolution for improving visual coherence while maintaining diversity.
\end{enumerate}

\section{Related work}

\subsection{High-fidelity image synthesis}
Recently developed GAN models apply the progressive growing approach to address the challenges, such as training instability and image quality degradation, in high-resolution image synthesis \cite{PGGAN, StyleGANv1, StyleGANv2, StackGAN, StackGAN++, HDGAN, LAPGAN, SinGAN, ConSinGAN, HP-VAE-GAN}. They also apply various techniques to improve fidelity. StackGAN \cite{StackGAN}, StackGAN++ \cite{StackGAN++}, and HDGAN \cite{HDGAN} apply a hierarchy of multi-level generators and discriminators to learn feature distributions at various scales. LAPGAN \cite{LAPGAN}, SinGAN \cite{SinGAN}, and ConSinGAN \cite{ConSinGAN} apply the Laplacian pyramid framework that increases the resolution of the image by upsampling and complements details at each level. StyleGAN \cite{StyleGANv1} and StyleGAN2 \cite{StyleGANv2} synthesize high-resolution images by adjusting the `style' at each scale while increasing the size of the feature maps. In addition, various methods such as minibatch discrimination \cite{PGGAN}, joint conditional-unconditional loss \cite{StackGAN++}, matching pair loss \cite{HDGAN}, mixing regularization \cite{StyleGANv1}, and path length regularization \cite{StyleGANv2} are employed to improve learning stability, image quality, and diversity.

Another way to improve the quality and resolution of image synthesis models is to apply image restoration or image fusion techniques. Widely used image processing techniques include deblurring, denoising, de-hazing, de-raining \cite{efficient-derain}, infrared and visible image fusion \cite{unsuperviesd_IVIF, IVIF_SOD}, and super-resolution \cite{ESRGAN, SwinIR, BSRGAN, ZSSR, MZSR}.
In this study, we apply super-resolution to synthesize high-quality UHR images. 

\begin{figure*}[h] 
\begin{center}
\centerline{\includegraphics[width=\textwidth]{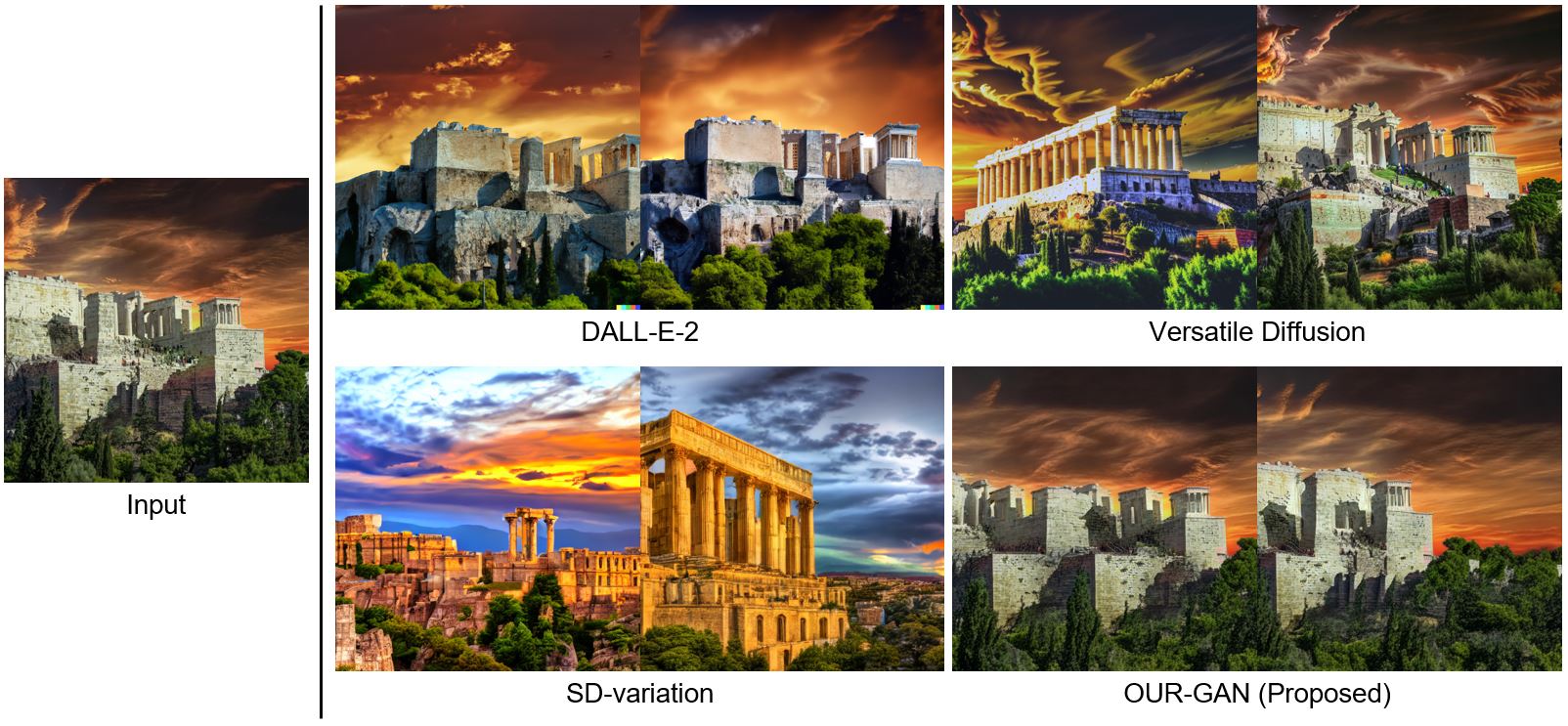}}
\caption{The output of DALL-E-2, Versatile Diffusion, SD-variation, and OUR-GAN at $512\times512$ resolution. The diffusion-based image variation models failed to reflect the texture, color, and lighting pattern of the input image, whereas OUR-GAN successfully represented them.}
\label{fig:image_variation}
\end{center}
\vskip -0.3in
\end{figure*}

\subsection{Diffusion models}

Diffusion models \cite{DPM, DDPM} generate images through an iterative denoising process. In recent years, diffusion models trained on large text-to-image datasets have exhibited outstanding quality and diversity in text-to-image conversion \cite{GLIDE, DALL-E-2, Imagen}. In addition, the latent diffusion model (LDM) \cite{LDM} significantly reduced the computational cost while maintaining the quality by conducting the denoising process in low-resolution latent space instead of image space.

Recent studies propose image manipulation methods based on pre-trained large-scale text-to-image diffusion models.
Kim et al. \cite{DiffusionCLIP} apply the CLIP loss \cite{CLIP} to fine-tune the denoising process for text-driven image manipulation.
Wu et al. \cite{UncoveringLatent}, Kwon et al. \cite{Asyrp}, and Jeong et al. \cite{Asyrp2} edit images by manipulating the latent space of the pre-trained diffusion models.
In \cite{P2P, NullText, PlugAndPlay}, image editing was performed by manipulating the cross-attention map without fine-tuning the pre-trained text-to-image model. 
Those three models were designed to preserve the layout of the reference image, and do not produce images with significantly different layouts.

Dreambooth \cite{Dreambooth} learns the embeddings of new concepts or objects from a small number of images, and then use these embeddings to synthesize images containing the concepts or objects in different contexts.
However, when training with a single training image, Dreambooth is prone to overfitting, often suffers from training instability, and is sensitive to the choice of hyperparameters \cite{ExntededTI}.

A few diffusion models take an image as input and generate different images with content and style similar to those of the input image. DALL-E-2 \cite{DALL-E-2} and Versatile Diffusion \cite{VersatileDiffusion} generate diverse image variations while maintaining the semantic content, style, and visual quality of the reference image. SD-variation \cite{SD-variation} extends Stable Diffusion \cite{StableDiffusion} to accept CLIP image embeddings instead of text embeddings, which enables the creation of image variations using Stable Diffusion.

These models can be used for one-shot or zero-shot image synthesis. However, they have a few limitations in one-shot image synthesis.
First, they often fail to adequately reflect the content or style of the input image. \cref{fig:image_variation} displays the results of DALL-E-2, Versatile Diffusion, and SD-variation. In \cref{fig:image_variation}, the diffusion models generated images with similar contents to the input image but failed to represent the texture, color, and lighting patterns of the input image. Although it is possible to control similarity to the reference image with Versatile Diffusion, such adjustment requires careful tuning and often results in a loss of diversity. We believe that this limitation is due to the fact that these models rely too much on the knowledge learned from the pre-training data and lack a powerful adaptation algorithm, such as gradient-based learning, to sufficiently learn the characteristics of the input image. Although this problem could be mitigated by fine-tuning the models to the input image, it is challenging to fine-tune those models on a small number of training examples while avoiding overfitting.

\begin{figure*}[!t] 
\begin{center}
\centerline{\includegraphics[width=\textwidth]{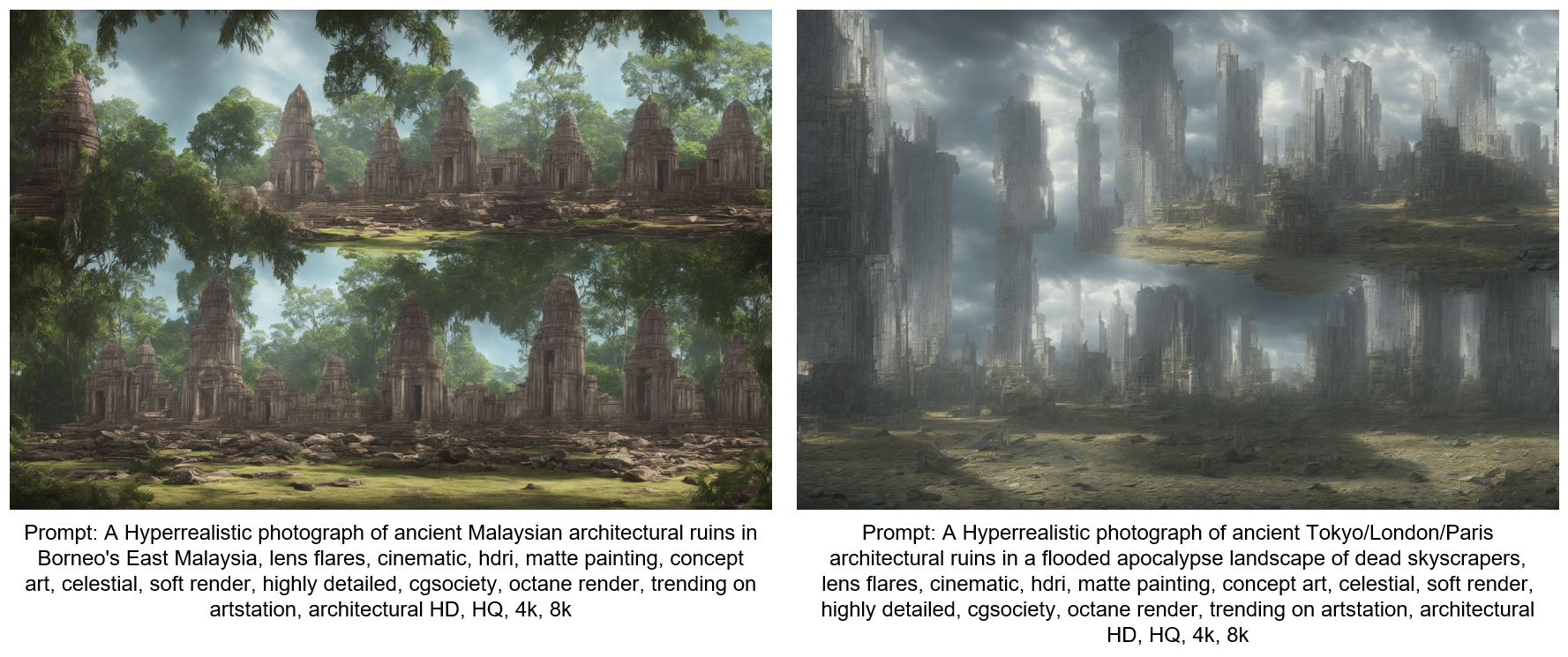}}
\caption{2K images synthesized by Stable Diffusion v2.1.r7 with prompts taken from the homepage of Stable Diffusion. These images have visually incoherent compositions of trees and ruins in the air.}
\label{fig:SD_2K}
\end{center}
\vskip -0.05in
\end{figure*}

Second, most diffusion models have limited output resolutions because they were trained with ordinary-sized images. For example, Stable Diffusion \cite{StableDiffusion} was trained at resolutions of $512\times512$ and $768\times768$ and cannot produce UHR images. Although it is possible to scale the output size of Stable Diffusion, the maximum output size on an RTX 3090 GPU with 24GB memory is limited to around 2K. Moreover, simply scaling the output size does not scale the contents and often results in visual incoherence, as shown in \cref{fig:SD_2K}.

\begin{figure}[!t] 
\begin{center}
\centerline{\includegraphics[width=0.9\columnwidth]{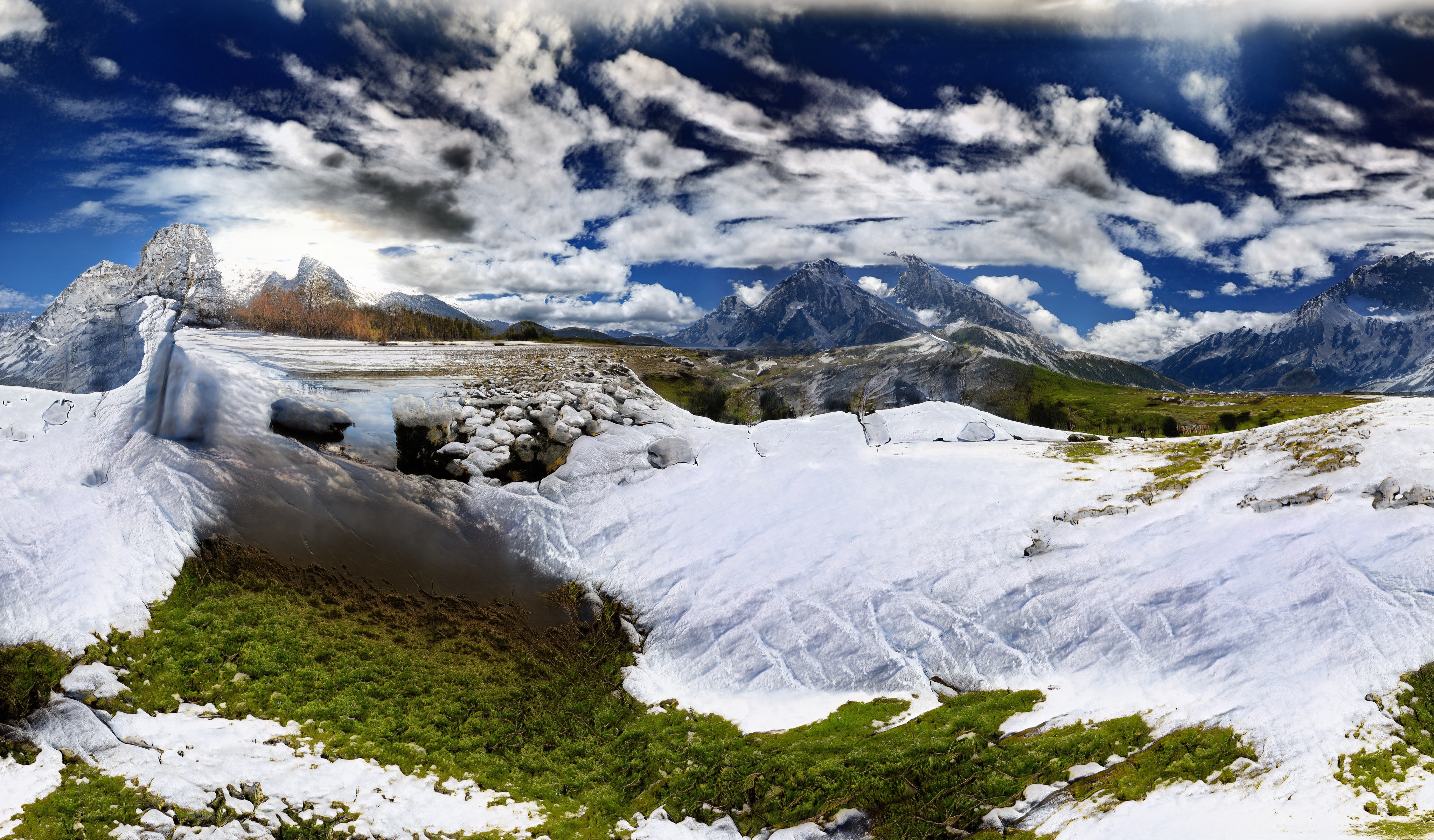}}
\caption{A 4K image synthesized by InfinityGAN \cite{InfinityGAN}}
\label{fig:InfinityGAN_sample}
\end{center}
\vskip -0.05in
\end{figure}

\subsection{Infinite-size image synthesis (IIS)}
\label{subsection:infinite_image_synthesis}

TextureNetwork \cite{TextureNetworks}, PSGAN \cite{PSGAN}, and InfiniteGAN \cite{inf-GAN} synthesize UHR images by connecting texture patches synthesized using existing methods. These models produce images composed of repeating textures but cannot synthesize complex and realistic images. There are a few prior studies on non-repetitive UHR image synthesis \cite{InfinityGAN, ALIS, TamingTransformer, NUWA-Infinity}. They synthesize images part by part and combine them to produce a full-size image. InfinityGAN, Taming Transformer, and ALIS \cite{ALIS} maintain the visual coherence of global shape through shared latent vectors.
InfinityGAN refers to a shared global latent vector while synthesizing each subregion. ALIS synthesizes partial images that smoothly connect them with each other by sharing latent anchor codes at regular distances in the coordinate system. To synthesize image parts, ALIS interpolates the anchor codes by spatially aligned adaptive instance normalization. Taming Transformer creates latent code maps in an autoregressive manner and then synthesizes subregion images conditioned on the latent codes at the corresponding coordinates.
These models can synthesize images of arbitrary size from relatively small training images. However, they are not one-shot synthesis models and require a large volume of training data. Moreover, because they do not learn from real UHR images, they cannot synthesize large shapes with fine details and long-range coherence, as shown in \cref{fig:InfinityGAN_sample}.

There is a one-shot infinite-size image synthesis model \cite{NUWA-Infinity} published after the preprint version of this paper \cite{OUR-GAN_arXiv}. However, NUWA-Infinity \cite{NUWA-Infinity} is composed of a large autoregressive model with 809M parameters and requires a large amount of computation and memory. In contrast, OUR-GAN consists of a significantly smaller network with only 23.4M parameters, which is merely 2.89\% of \cite{NUWA-Infinity}. Moreover, a larger version of our model, OUR-GAN-16K, can synthesize 16K images with 35.1M parameters and is trainable on a single consumer GPU.

\begin{figure*}[ht] 
\vskip 0.1in
\begin{center}
\centerline{\includegraphics[width=\textwidth]{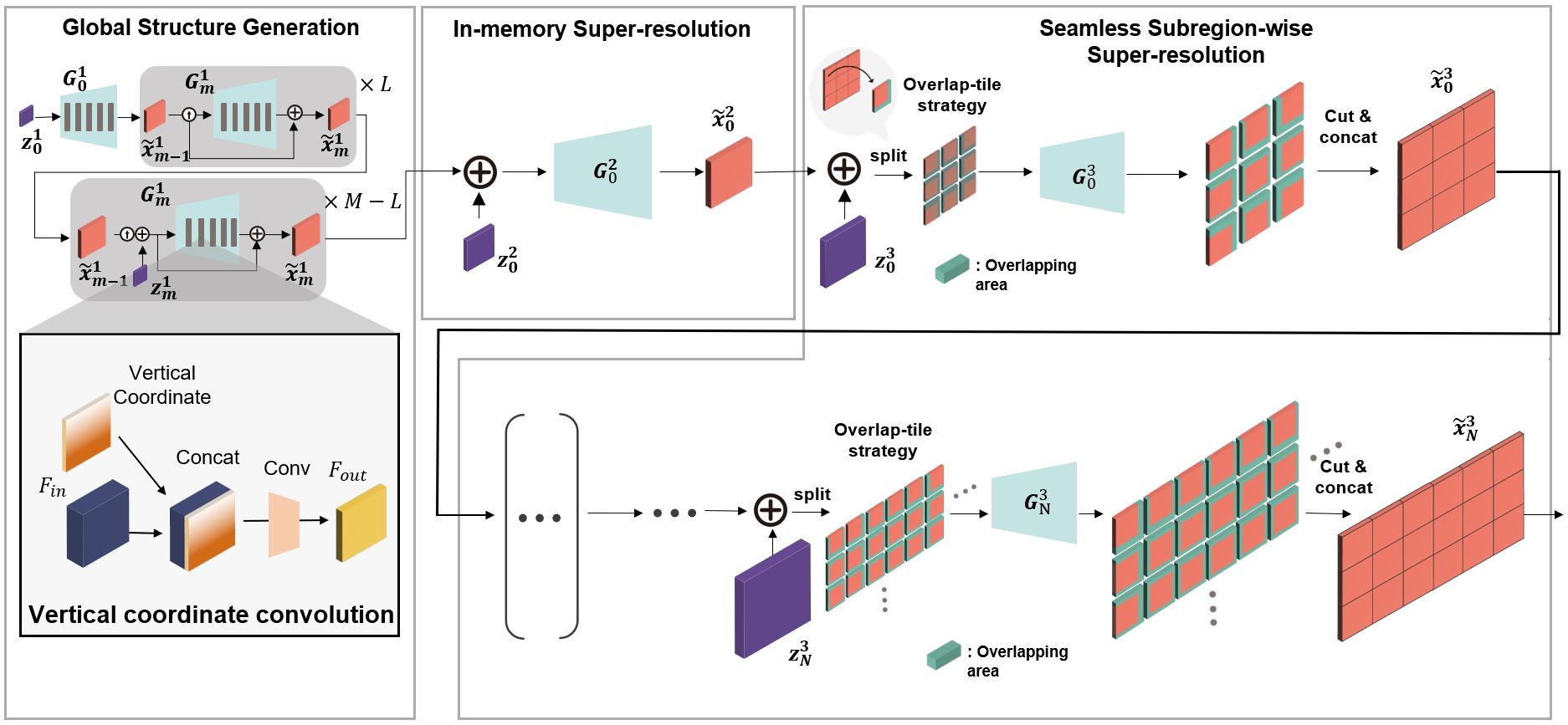}}
\caption{One-shot UHR image synthesis framework. OUR-GAN synthesizes UHR images in three or more steps: 1) global structure generation to synthesize diverse and visually coherent initial images at low resolution, 2) in-memory super-resolution to scale up the synthesized image within the memory limit, and 3) one or more steps of seamless subregion-wise super-resolution to produce UHR images with limited memory.
}
\label{fig:Synthesis_procedure}
\end{center}
\vskip -0.2in
\end{figure*}

\subsection{One-shot image synthesis}
\label{subsection:oneshot_image_synthesis}
One-shot learning, which is a special case of few-shot learning  \cite{tang2022, zha2023}, is the task of training a model with a single training example.
There is a large volume of previous work on one-shot learning for computer vision tasks, including object image recognition, face recognition, medical diagnosis, instance segmentation, semantic segmentation, and object detection.
However, it is relatively recent that one-shot generative models have been actively studied \cite{SinGAN, ConSinGAN, InGAN, HP-VAE-GAN, GPNN, GPDM, ExSinGAN, PetsGAN, SinFusion, SinDiffusion, SinDDM, SIV-GAN, IMAGINE}.
InGAN \cite{InGAN}, SinGAN \cite{SinGAN}, ConSinGAN \cite{ConSinGAN}, and HP-VAE-GAN \cite{HP-VAE-GAN} learn the internal patch distribution of the training image and then synthesize new images based on the patch distribution. 
InGAN \cite{InGAN} is the first conditional one-shot GAN model that generates natural images. The generator of InGAN learns patch distribution at multiple scales guided by a multi-scale patch discriminator and a cycle-consistent loss. In synthesis, it generates images that have the same patch distribution but with different sizes and shapes. SinGAN \cite{SinGAN} learns the distribution of patches at multiple scales by a hierarchy of multi-level generators and discriminators. SinGAN prevents overfitting by applying the Laplacian pyramid framework (LAPGAN) \cite{LAPGAN}. In this case, each generator is trained one at a time, freezing the previous generators. HP-VAE-GAN \cite{HP-VAE-GAN} improves diversity and overcomes the mode collapse problem of GAN-based models by integrating PatchVAE \cite{PatchVAE} with GAN. SIV-GAN \cite{SIV-GAN} improves global shape coherence and reduces overfitting by discriminating the image in the content and layout branches separately. However, the aforementioned one-shot synthesis models have room for improvement in terms of global shape coherence and shape diversity.

A few recent studies synthesize images by inverting a pre-trained GAN or classifier \cite{IMAGINE, ExSinGAN, PetsGAN}. There also exist one-shot image synthesis models not based on GAN. GPNN \cite{GPNN} and GPDM \cite{GPDM} adapt the classical patch-based method that does not require any training. They improve visual quality by directly using the patches of the training image, but cannot synthesize novel patches. Nikankin et al. \cite{SinFusion}, Wang et al. \cite{SinDiffusion}, and Kulikov et al. \cite{SinDDM} apply denoising diffusion models \cite{DPM, DDPM} to one-shot image synthesis.

However, none of the above studies present a synthesized UHR image with a resolution higher than 4K. In theory, it is possible to extend an existing one-shot synthesis model to synthesize UHR images. However, in practice, simply extending those models for UHR image synthesis requires a large amount of GPU memory.

\section{One-shot Ultra-high-Resolution Generative Adversarial Networks}
\subsection{Overview of OUR-GAN framework}
OUR-GAN synthesizes UHR images with limited GPU memory in three or four steps, as illustrated in \cref{fig:Synthesis_procedure}. In the first step, OUR-GAN generates the global structure of the image at low resolution. Then, in the second step, it increases the resolution to as high as possible within the memory limit through in-memory super-resolution. 
In the third step, it produces a UHR image by further increasing the resolution beyond the memory limit through seamless subregion-wise super-resolution. When synthesizing UHR images with resolutions higher than 4K, as in OUR-GAN-16K described in \cref{result_16K}, we apply an additional subregion-wise super-resolution step to further increase the resolution.

OUR-GAN is an extension of one-shot synthesis models designed to generate UHR images with limited memory and computation. Therefore, in contrast to the IIS models in \cref{subsection:infinite_image_synthesis}, OUR-GAN does not require a large volume of training data and can be trained with only a single training image. Moreover, because it learns both global structures and fine details from a real UHR image, it can generate not only large-sized images but also images containing large shapes with fine details, as Figures \ref{fig:16K_Image_Synthesis}, whereas the conventional IIS models rely on the patch distribution learned from relatively small-sized images and are difficult to synthesize large shapes, as shown in \cref{fig:InfinityGAN_sample}. Because it adapts to the input image through powerful gradient-based learning, it can reflect the content and style of the input image better than the diffusion models that rely heavily on pre-trained knowledge. In addition, OUR-GAN exhibits improved visual coherence by exploiting the correlation between visual components and their vertical coordinates. Compared with the one-shot synthesis models in \cref{subsection:oneshot_image_synthesis}, OUR-GAN synthesizes images with a significantly higher resolution and improved visual coherence.

One disadvantage of super-resolution compared with the patch distribution-based IIS models is that the output resolution is restricted by that of the training image. In spite of that, OUR-GAN can synthesize high-quality images with $2\times$ $\sim$ $4\times$ higher resolution than the training image by exploiting the internal recurrence of information, similar to \cite{ZSSR} and \cite{MZSR}. \cref{fig:16K_Image_Synthesis} displays a 16K image synthesized by OUR-GAN trained with an 8K image. Our demo page presents more samples with various resolutions synthesized by OUR-GAN.

\begin{figure*}[h] 
\begin{center}
\centerline{\includegraphics[width=\textwidth]{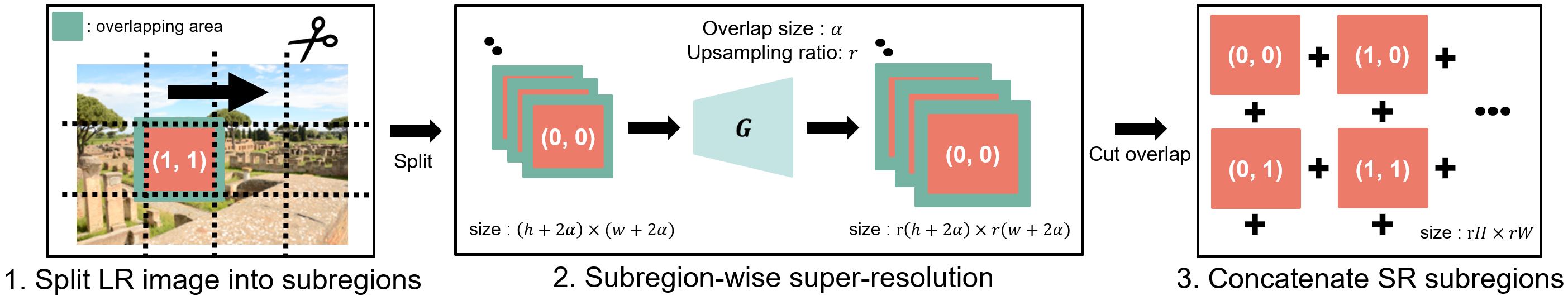}}
\caption{Seamless subregion-wise super-resolution. OUR-GAN first splits the input image into small overlapping ubregions. Then it increases the resolution of each subregion including the overlapping area. Finally, it removes the overlapping area from the super-resolved subregion images and concatenates them to create a seamless full-size image. OUR-GAN minimizes additional memory requirements by estimating overlap size to prevent discontinuity from the effective receptive field of the super-resolution model.}
\label{fig:Subregion-wise_SR}
\end{center}
\vskip -0.25in
\end{figure*}

\begin{figure}[t] 
\begin{center}
\centerline{\includegraphics[width=0.8\columnwidth]{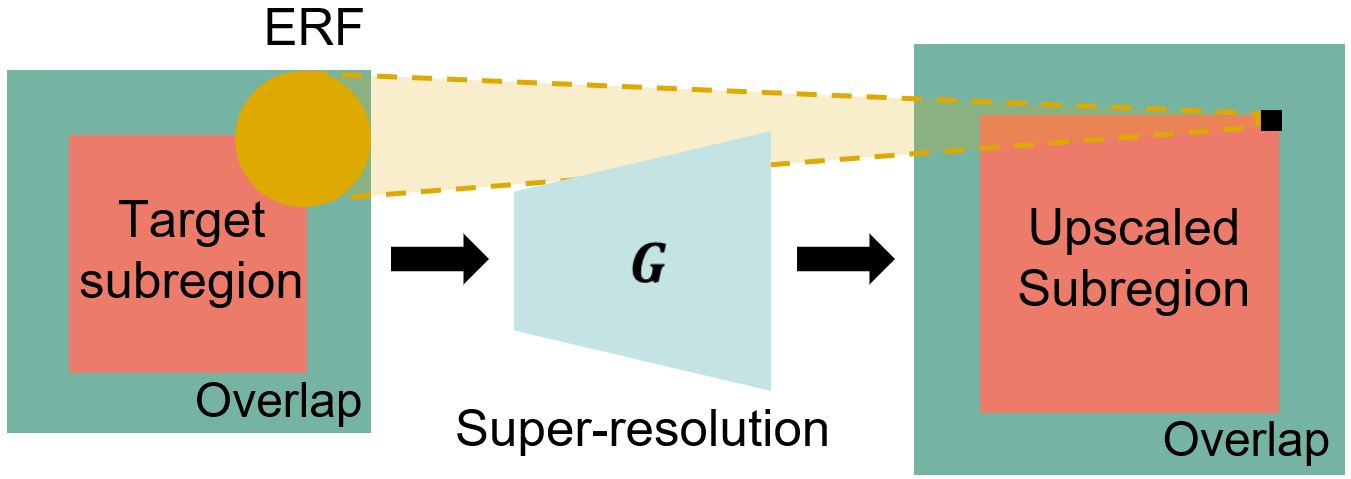}}
\caption{Removing information loss by overlapping subregions by the ERF radius.
If the overlap is larger than the radius of ERF, the super-resolution model up-scales the target subregion by referring same input when synthesized without subregion.}
\label{fig:ERF_overlap}
\end{center}
\vskip -0.25in
\end{figure}

\subsection{Global structure generation}
\label{subsection:global_structure_generation}

It is challenging to synthesize diverse images with globally coherent shapes from a single training image. We compared multiple one-shot synthesis models \cite{SinGAN, ConSinGAN, HP-VAE-GAN, SIV-GAN, PetsGAN, SinFusion, SinDDM} in a preliminary experiment and selected HP-VAE-GAN \cite{HP-VAE-GAN} as the baseline model because it exhibited higher diversity than other models. However, HP-VAE-GAN often produced visually incoherent images. Therefore, we improved the visual coherence of the baseline model by applying vertical coordinate convolution \cite{CoordConv}. HP-VAE-GAN synthesizes images through a hierarchical patch-based generation scheme as Eq. (1)-(3), where $G^1_m$, $\tilde{x}^1_m$ and $z^1_m$ denote the generator, synthesized image, and Gaussian noise vector at scale $m$, respectively. The superscript `$1$' indicates the first step of the framework shown in \cref{fig:Synthesis_procedure}, and the symbol $\uparrow$ represents upsampling.

\begin{numcases}{\tilde{x}^1_m=}
   G^1_0(z^1_0) &  \label{eq:1st_step_infer_1}
   \\ \uparrow\tilde{x}^1_{m-1} + G^1_m(\uparrow\tilde{x}^1_{m-1}) & $1 \leq m \leq L$ \label{eq:1st_step_infer_2} 
   \\ \uparrow\tilde{x}^1_{m-1} + G^1_m(\uparrow\tilde{x}^1_{m-1}+z^1_m) & $L < m \leq M$ \label{eq:1st_step_infer_3},
\end{numcases}

HP-VAE-GAN first synthesizes an initial image from Gaussian noise $z^1_0$, as shown in Eq. ({\ref{eq:1st_step_infer_1}}), and then gradually increases the resolution by the Laplacian framework, as shown in Eq. ({\ref{eq:1st_step_infer_2}}) and ({\ref{eq:1st_step_infer_3}}). In the early stages of $1 \leq m \leq L$, it applies patchVAE \cite{PatchVAE}, as shown in Eq. ({\ref{eq:1st_step_infer_2}}), because the diversity of GAN models is limited due to the mode collapse problem. In contrast, in the late stages of $L < m \leq M$, it applies patchGAN \cite{PatchGAN}, as Eq. ({\ref{eq:1st_step_infer_3}}), which synthesizes images with better detail fidelity than VAE.

\par

While HP-VAE-GAN generates diverse images, it often produces visually incoherent images in which the vertical arrangement of visual elements is inadequate. We alleviated this issue by exploiting the spatial bias of the scenery image.
The vertical coordinates of the visual components in the scenery image have a strong bias \cite{HANet}. Changes in the vertical position of components in the landscape image can result in a visually incoherent or unreasonable layout, such as the sky below a mountain, whereas changes in the horizontal position rarely cause serious problems. Therefore, we extended the HP-VAE-GAN to synthesize the visual elements of an image by reflecting their vertical position.

The coordinate convolution \cite{CoordConv} explicitly utilizes position information by concatenating coordinate channels to the input feature maps. The coordinate channels contain the horizontal and vertical coordinates of each location normalized to range from -1 to 1. Using coordinate channels, the model can learn the relation between visual elements and their locations. However, when the model learns from a single image, coordinate convolution causes a severe loss of diversity because the coordinate channels associate the visual elements too strongly with their absolute coordinates. Therefore, we only concatenate the vertical coordinate channel and do not concatenate the horizontal coordinate channel, as \cref{fig:Synthesis_procedure}. The absence of the horizontal coordinate channel allows the model to produce a variety of layouts because it attenuates the correlation between visual elements and their location. We apply the vertical coordinate convolution to all convolution layers in the first step, which generates the global structure. However, we do not apply it to the subsequent steps because they only increase the resolution through super-resolution.
A prior study \cite{CoordConv} also utilizes vertical coordinates to obtain attention value, but their method significantly differs from ours.
\par

\begin{figure*}[ht] 
\vskip -0.1in
\begin{center}
\centerline{\includegraphics[width=0.80\linewidth]{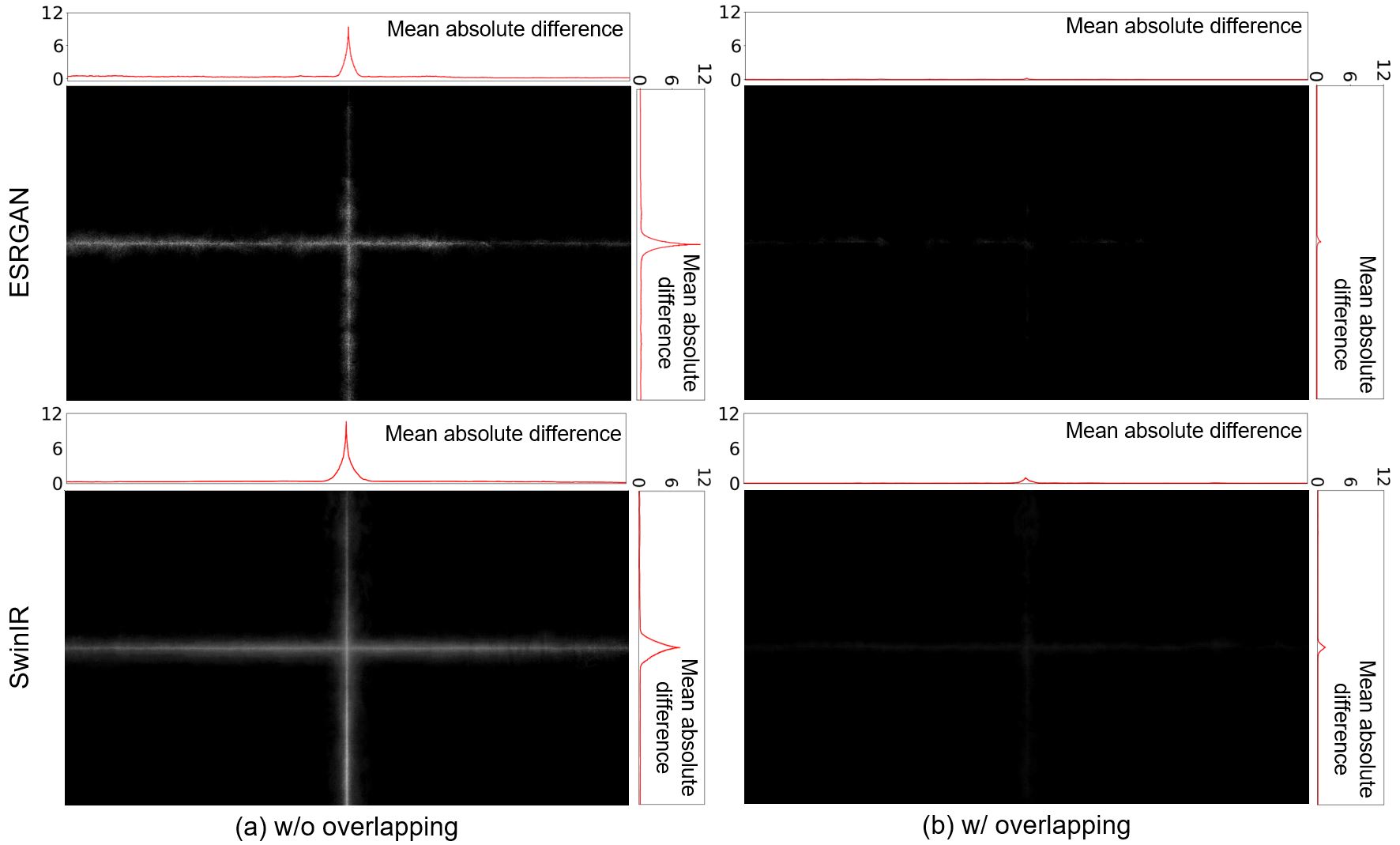}}
\caption{The absolute difference between images produced by subregion-wise super-resolution and by ordinary super-resolution which increases resolution in one step without splitting the input image. (a) and (b) are the difference maps of 4K images synthesized with and without overlapping, respectively. The brighter the color, the more significant the difference. The overlap size is 10 pixels for ESRGAN and 24 pixels for SwinIR.}
\label{fig:Border_heatmap}
\end{center}
\vskip -0.35in
\end{figure*}

\subsection{In-memory and subregion-wise super-resolution}

With the initial image generated in the first step, OUR-GAN increases the resolution of the synthesized image through super-resolution in the following steps. Whereas the first step focuses on diversity and visual coherence, the super-resolution steps focus on the fidelity of detail.
In particular, the third and fourth steps apply subregion-wise super-resolution to produce UHR images with limited memory.

The greatest technical challenge in these steps is learning super-resolution with a single training image.
We overcome this challenge through the pre-training and fine-tuning strategy.
We also considered zero-shot super-resolution models \cite{ZSSR, MZSR} that do not require a large volume of training data.
However, we obtained better image quality than the zero-shot models when we pre-trained ordinary super-resolution models, such as ESRGAN \cite{ESRGAN} and SwinIR \cite{SwinIR}, on public datasets and then fine-tuned them on the training image.


In the second step, we add random noise $z^2_0$ to the previously synthesized image $\tilde{x}^1_M$, then increase the resolution by a super-resolution model $G^2_0$ as $\tilde{x}^2_0 = G^2_0(\tilde{x}^1_M + z^2_0)$.
In the third and fourth steps, we divide the low-resolution image into subregions, apply super-resolution to each of the subregion images, and then concatenate the up-scaled subregions images into a single high-resolution image, as shown in \cref{fig:Subregion-wise_SR}.

Without careful design, subregion-wise super-resolution leads to discontinuity at the boundaries between subregions. In general, the contexts on the opposite side of the boundary are replaced by zero values. Therefore, when generating each subregion, the model does not refer to the input feature maps of neighboring subregions. As a result, the output pixels located on the opposite side of the boundary may differ excessively because, in spite of the close distance, they are generated from different contexts. This issue is common for subregion-wise super-resolution models based on CNN and Swin Transformer \cite{SwinTransformer}.

There are a few prior studies that address this issue \cite{UNet, InfinityGAN}. U-Net \cite{UNet} applies an overlap-tile strategy for image segmentation that expands the input subregion area to remove the influence of the zero values at the boundary. In \cite{InfinityGAN}, InfinityGAN applies a carefully designed network to avoid zero-padding, which alternates convolutions and transpose convolutions. As the latter requires a specially designed network, we applied the former and improved it to minimize memory requirements.

In \cite{UNet}, U-Net overlaps the tiles to a substantial length to prevent discontinuity at the boundary. In theory, the overlap should be larger than the receptive field of the layer. The size of the theoretical receptive field (TRF) linearly increases with the depth of the network, which is large for a deep convolutional neural network (CNN) and Swin Transformer \cite{ERF, DiNA}. In contrast, \cite{ERF} analyzed the effective receptive field (ERF) of deep CNNs and reported that ERF is significantly smaller than TRF and increases with the square root of network depth. Inspired by \cite{ERF}, we set the overlap size according to the radius of ERF, as shown in \cref{fig:ERF_overlap}. Following \cite{ERF}, we measured the ERF of the super-resolution models by the gradient of an output pixel with respect to the input pixels. Based on the results, we chose 10 and 24 pixels for the overlap size of the subregion-wise ESRGAN and SwinIR, respectively. The two values are the radius of the circles that cover approximately 98\% of the gradient magnitude. Nevertheless, they are less than 7.4\% of the TRF of the models. OUR-GAN prevents discontinuity at the subregion boundaries with a minimal amount of additional memory, as shown in  \cref{fig:Border_heatmap}.

\begin{figure*}[h] 
\begin{center}
\centerline{\includegraphics[width=0.9\linewidth]{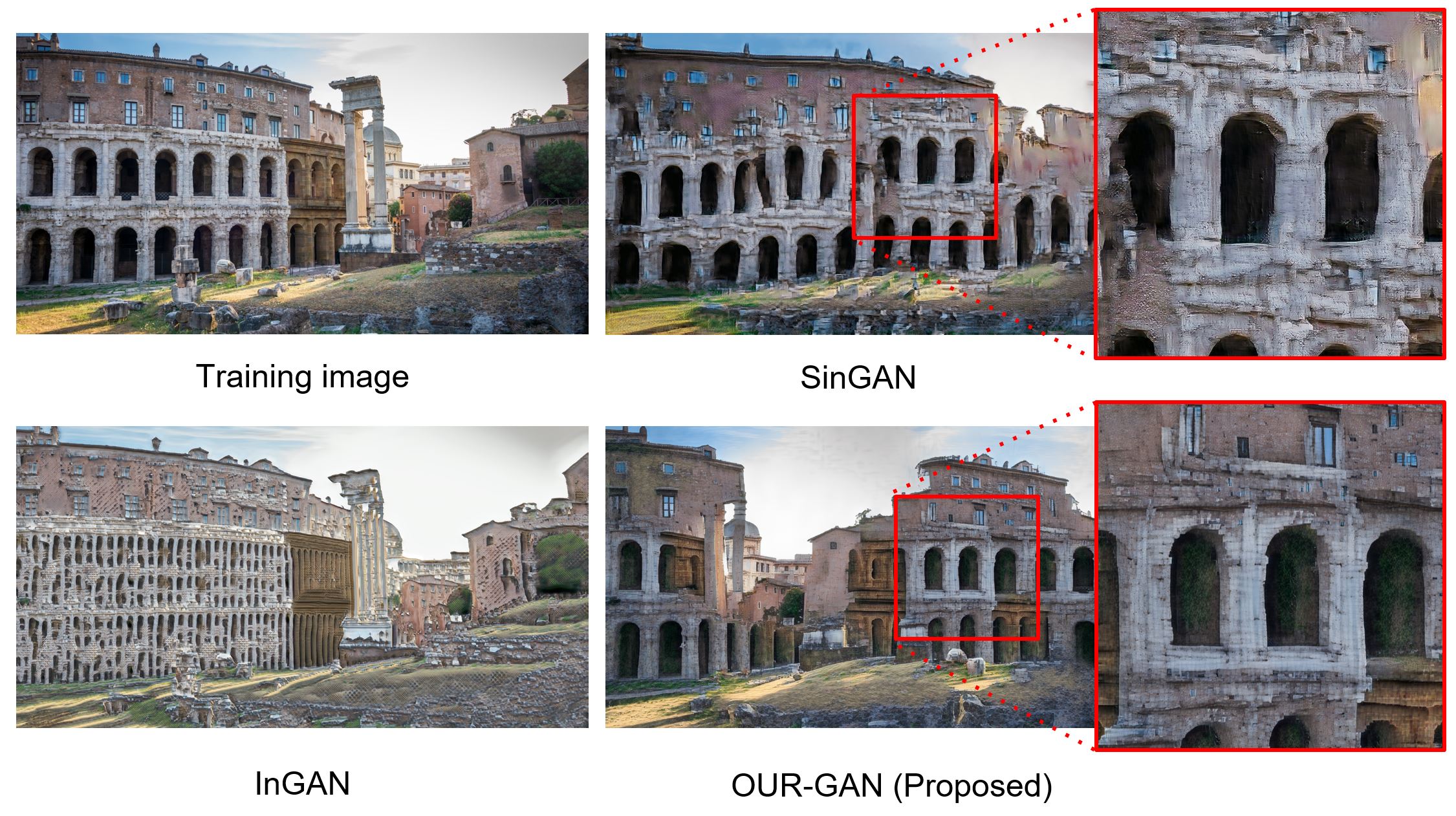}}
\caption{Qualitative comparison of non-repetitive 4K Image Synthesis. OUR-GAN synthesized globally coherent structures, while InGAN generated distorted objects.
SinGAN also produced globally coherent structures, but inferior to OUR-GAN in detail quality.}
\label{fig:4K_scenery_synthesis}
\end{center}
\end{figure*}

With an overlap $\alpha$ set to the ERF of the super-resolution model, the detailed procedure of the subregion-wise super-resolution is as follows.
First, we add random noise ${z}^3_0$ to the previously synthesized image $\tilde{x}^2_{0} \in \mathbb{R}^{3\times H \times W }$ and split it into $K$ overlapping subregions $s \in \mathbb{R}^{3 \times (h+2\alpha)\times (w+2\alpha)}$ as
\begin{equation}
    {\{s_1, s_2, ..., s_k \} = split_{hw}(\tilde{x}^2_{0} + z^3_{0})},
\end{equation}
where $K:=\lceil {H \over h} \rceil \times \lceil {W\over w} \rceil$ denotes the number of subregions and $\alpha$ denotes overlap size.
The subregion size $(h + 2\alpha) \times (w + 2\alpha)$ should not exceed the maximum size that super-resolution can be applied within the GPU memory limit.

Then, the super-resolution model $G_0^3$ upscales each subregion one by one as
\begin{equation}
    { \hat{s}_i = G_{0}^3(s_i) \in \mathbb{R}^{3\times r(h+2\alpha) \times r(w+2\alpha)}},
\end{equation}
where $i = {1,2,...,K}$ is index of subregions, $r$ denotes upscale ratio and $\hat{s}_i$ denotes upscaled subregion. 


Third, we remove overlapping area from each upscaled subregion images as 
\begin{equation}
    { \Acute{s}_i = trim_{r\alpha}(\hat{s}_i) \in \mathbb{R}^{3 \times rh\times rw}},
\end{equation}
where $trim_{b}(\cdot)$ removes $b$ boundary rows and columns from the input feature map.

Finally, we concatenate them to create a seamless full-size image as
\begin{equation}
    { \tilde{x}_0^3 = concat(\Acute{s}_1,\Acute{s}_2,...,\Acute{s}_k) \in \mathbb{R}^{3 \times rH \times rW} },
\end{equation}
where $\Acute{s}_i$ is the upscaled subregion with the scaled overlap removed, and $\tilde{x}_0^3$ is the output of third step.
Such a subregion-wise super-resolution can be repeated multiple times, but, in this research, we applied only twice at maximum, which was sufficient to generate 16K UHR images.

The memory complexity of the subregion-wise super-resolution is proportional to its output resolution as in \cref{eq:memory-complexity}.
\begin{equation}
\begin{split}
    M(h, w, r, \alpha) &\propto r(h+2\alpha) \times r(w+2\alpha)   \\
            &= r^2hw + 4r^2(h+w)\alpha + 4r^2\alpha^2,
\label{eq:memory-complexity}
\end{split} 
\end{equation}
where $4r^2(h+w)\alpha + 4r^2\alpha^2$ is the overhead for the overlap. For example, the TRF and ERF of our SwinIR are 324 and 24, respectively. When $h = 270$, $w = 512$, and $r = 4$, 
the value of \cref{eq:memory-complexity} is 12.98M with $\alpha$ set to TRF, but only 2.55M with $\alpha$ set to ERF. The latter is $5.09 \times$ smaller than the former.

To compare the results of subregion-wise super-resolution with and without overlap, we measured the absolute difference between the high-resolution images produced by the subregion-wise super-resolution and by the ordinary super-resolution. \cref{fig:Border_heatmap} displays the results. In \cref{fig:Border_heatmap}, the subregion-wise super-resolution without overlap exhibits a significant discrepancy from the result of ordinary convolution at the subregion boundaries. However, the small size of overlap successfully attenuates the discrepancy. These results suggest that such small overlap sizes are sufficient to suppress the discontinuity to an invisible level for both the CNN-based and Swin Transformer-based models.

\begin{figure*}[h] 
\vskip 0.1in
\begin{center}
\centerline{\includegraphics[width=0.95\textwidth]{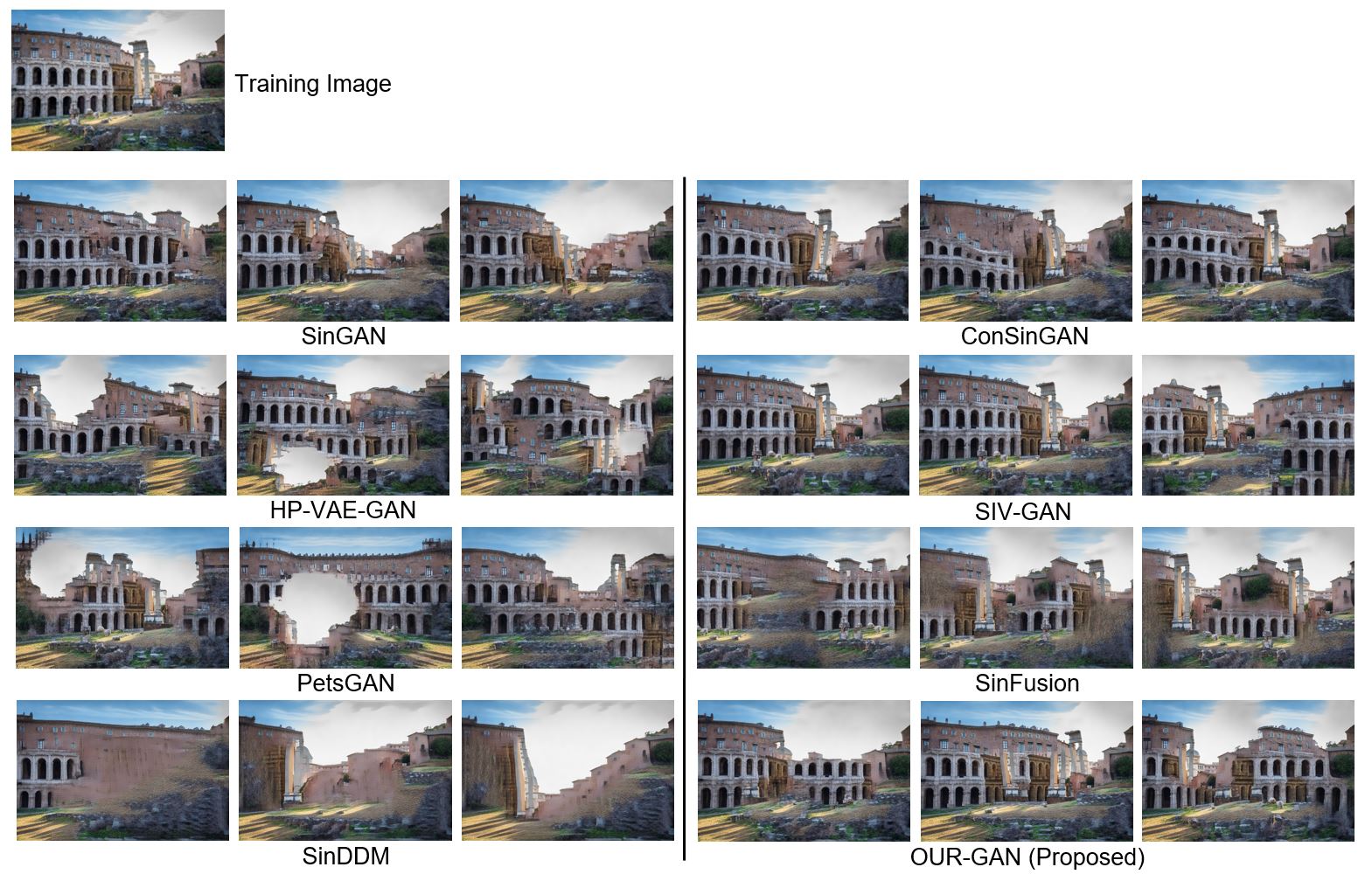}}
\caption{4K images synthesized by OUR-GAN and seven one-shot synthesis models: SinGAN \cite{SinGAN}, ConSinGAN \cite{ConSinGAN}, HP-VAE-GAN \cite{HP-VAE-GAN}, SIV-GAN \cite{SIV-GAN}, PetsGAN \cite{PetsGAN}, SinFusion \cite{SinFusion} and SinDDM \cite{SinDDM}. While the baseline models exhibited limited diversity or produced visually incoherent images, OUR-GAN synthesized visually coherent and diverse images.
}
\label{fig:Comparison_global}
\end{center}
\vskip -0.3in
\end{figure*}

\section{Experiments}
\subsection{Experimental settings}

\textbf{Datasets and experimental environment}
There are a few datasets that contain UHR images with 4K or higher resolutions \cite{RAISE, UHDSR4K, DIV8K}. However, they also include low-resolution images or low-quality images with out-of-focus, light scattering, noise, etc. To evaluate 4K image synthesis models, we used two datasets solely composed of high-quality UHR images. The first dataset is a subset of the RAISE dataset only composed of high-quality UHR images. RAISE is a public image dataset primarily designed for the evaluation of digital forgery detection \cite{RAISE}. We randomly selected 50 UHR scenery images from the RAISE dataset and then repeated the process of replacing low-quality images with other randomly selected UHR images. \cref{appendix:RAISE} displays the thumbnails of all selected images.

For a more reliable evaluation, we also collected a new UHR image dataset, \textbf{S}cenery and \textbf{T}exture-\textbf{4K} (\textbf{ST4K}). The ST4K dataset consists of a total of 50 copyright-free images collected from the Internet with a minimum resolution of $4,096\times2,160$ pixels. ST4K includes 25 diverse natural and urban scenery images containing multiple global and structural patterns, as well as 25 high-quality texture images. \cref{appendix:ST4K} presents the thumbnails of ST4K samples. The full-size images are downloadable from our demo page. To focus on the performance of non-repetitive image synthesis, we only used the scenery images in ST4K in the evaluation.

\begin{table}[t!]
\begin{center}
\begin{footnotesize}
\begin{sc}
\renewcommand{\tabcolsep}{0.7mm}
\begin{tabular}{lcccc}
\toprule
steps          & 1st & 2nd   & 3rd   & 4th  \\
\midrule
OUR-GAN       & 256x135 & 1,024x540 & 4,096x2,160 & -  \\
OUR-GAN-16K   & 256x135 & 1,024x540 & 4,096x2,160 & 16,384x8,640  \\
\bottomrule
\end{tabular}
\end{sc}
\end{footnotesize}
\end{center}
\vskip -0.1in
\caption{The resolution of the synthesized image at each step. As the aspect ratio of each image is different, we only denote the maximum vertical resolution of the synthesized image.}
\label{tab:4K_16K_image_resolution}
\end{table}

We used ESRGAN and SwinIR as the baseline of our super-resolution model. 
The algorithms, structures and hyperparameters of OUR-GAN are presented in \cref{appendix:training}, \ref{appendix:structure} and \ref{appendix:hyperparam}.
To evaluate one-shot generative models, we trained a separate model for each training image. We trained each model on a single GPU: RTX-2080 and GTX-1080 with 8 GB memory for 4K image synthesis and RTX-3090 with 24 GB memory for 16K image synthesis.
The output resolution of each step is displayed in \cref{tab:4K_16K_image_resolution}.

\textbf{Evaluation metrics}
In experiments, we measured Single Image Fréchet Inception Distance (SIFID) \cite{SinGAN} to quantify the quality of synthesized images and Learned Perceptual Image Patch Similarity (LPIPS) \cite{LPIPS} to quantify the diversity of synthetic patterns as \cite{SIV-GAN}.
Prior studies measure SIFID at different levels of the Inception network. \cite{SinGAN} measures SIFID from low-level feature maps extracted from an early layer, whereas \cite{SIV-GAN} measures from mid-level feature maps retrieved from the convolution layer just before the auxiliary classifier. We used both methods to measure the quality of both local and global patterns.
All results presented in this paper are averaged measurements over 50 generated samples, except for InGAN \cite{InGAN}, which is a conditional model that generates one output image from the input image, not from noise. 
\par

\subsection{One-shot 4K image synthesis}

We first evaluated the performance of one-shot image synthesis.
In this experiment, we compared the image produced by OUR-GAN with two baseline one-shot synthesis models that can synthesize non-repetitive high-resolution images on a single consumer GPU: InGAN \cite{InGAN} and SinGAN \cite{SinGAN}. 
One-shot synthesis models have limited output resolution; none of these studies present a synthesized UHR image with a resolution higher than 4K. Although SinGAN and InGAN did not present a 4K resolution image, we selected them as baseline models. 
Only InGAN, SinGAN and PetsGAN proposed methods to increase the output resolution using their models. However, we excluded PetsGAN from the baseline models because the maximum output resolution with 24 GB of memory is lower than 1K.
InGAN \cite{InGAN} synthesizes images by taking the training image as input and up-scaling the image to maximum resolution through a geometric transformation layer. SinGAN \cite{SinGAN} synthesizes a low-resolution initial image and then increases the resolution through the Laplacian framework.

\begin{table}[t!]
\begin{center}
\begin{small}
\begin{sc}
\renewcommand{\tabcolsep}{2.5mm}
\begin{tabular}{lcccc}
\toprule
            & \multicolumn{2}{l}{ Empirical } & \multicolumn{2}{l}{ Theoretical }  \\
            & \multicolumn{2}{l}{ measurement } & \multicolumn{2}{l}{ estimation }  \\
            & train & synthesis & train & synthesis  \\
\midrule
InGAN     & 22.29 & 21.32 & 13.82  & 9.08 \\
SinGAN     & 22.46 & 22.46 & 10.40  & 9.16 \\
\midrule
OUR-GAN   & \textbf{5.59} & \textbf{4.29} & \textbf{4.34} & \textbf{1.35} \\
\bottomrule
\end{tabular}
\end{sc}
\end{small}
\end{center}
\vskip -0.1in
\caption{Maximum GPU memory consumption (GB) of one-shot synthesis models for training and synthesis.} 
\label{tab:memory_efficiency}
\end{table}

We conducted a comparative evaluation at 4K resolution because NVIDIA's consumer GPUs, the RTX series, have a maximum memory size of 24 GB, limiting the output resolution of the baseline models to 4K. Moreover, the maximum output resolution of the baseline models that can be trained with 24 GB of memory is limited to only 1K. Therefore, we trained the baseline models at 1K resolution and then extended them to synthesize 4K images by applying the techniques presented in \cite{InGAN} and \cite{SinGAN}, respectively. The extended InGAN applies the geometric transform layer to generate a 4K image with the same internal patch distribution as the training image. The extended SinGAN comprises a 1K image synthesis module and a super-resolution module that increases the image resolution from 1K to 4K. In the training of the super-resolution module, we assigned a heavy weight of 100 to the reconstruction loss. To be fair, we also trained OUR-GAN at 1K resolution, although OUR-GAN can be trained to synthesize 4K images with less than 8 GB of GPU memory.

\begin{table}[t!]
\begin{center}
\begin{small}
\begin{sc}
\renewcommand{\tabcolsep}{8mm}
\begin{tabular}{lcc}
\toprule
          & \multicolumn{2}{c}{ST4K}  \\
          & \multicolumn{2}{c}{SIFID↓}  \\
Layer indices & 5th & 13th  \\
\midrule
InGAN     & 5.42 & 3.11   \\
SinGAN    & 8.84 & 3.82   \\
\midrule
OUR-GAN   & \textbf{2.94} & \textbf{2.29} \\
\bottomrule
\end{tabular}
\end{sc}
\end{small}
\end{center}
\vskip -0.1in
\caption{The SIFID of 4K non-repetitive images synthesized by one-shot synthesis models. We measured SIFID from the feature maps of the 5th and 13th convolution layers of the Inception network. ↓ indicates that the lower the better.
}
\label{tab:4K_image_synthesis}
\end{table}

\begin{table*}[!ht]
\begin{center}
\begin{small}
\begin{sc}
\renewcommand{\tabcolsep}{1.8mm}
\begin{tabular}{lcccccccccccc}
\toprule
               & \multicolumn{6}{c}{ST4K} & \multicolumn{6}{c}{RAISE} \\
               & \multicolumn{4}{c}{SIFID↓} & \multicolumn{2}{c}{LPIPS↑} & \multicolumn{4}{c}{SIFID↓} & \multicolumn{2}{c}{LPIPS↑} \\
               & \multicolumn{2}{c}{ESRGAN} & \multicolumn{2}{c}{SwinIR} & ESRGAN & SwinIR & \multicolumn{2}{c}{ESRGAN} & \multicolumn{2}{c}{SwinIR} & ESRGAN & SwinIR \\
Layer indices  & 5th & 13th & 5th & 13th &  & & 5th & 13th & 5th & 13th & \\
\midrule	
SinGAN      & 1.54 & 1.67 & 1.14  & 1.49 & 0.42          & 0.40          & 1.50 & 1.77 & 1.23 & 1.74 & 0.41 & 0.39 \\
ConSinGAN   & 1.45 & 1.58 & 1.03  & 1.42 & 0.41          & 0.39          & 1.36 & 1.67 & 1.13 & 1.68 & 0.40 & 0.38 \\
HP-VAE-GAN  & 1.47 & 1.56 & 1.05  & 1.38 & \textbf{0.53} & \textbf{0.52} & 1.34 & 1.62 & 1.11 & 1.60 & \textbf{0.48} & \textbf{0.47} \\
SIV-GAN     & 1.55 & 1.54 & 0.98  & 1.27 & 0.42          & 0.40          & 1.46 & 1.58 & 1.08 & 1.51 & 0.40 & 0.38 \\
PetsGAN     & 1.83 & 1.84 & 1.25  & 1.55 & 0.47          & 0.46          & 1.30 & 1.61 & 1.32 & 1.73 & 0.43 & 0.41 \\
SinFusion   & 2.39 & 1.95 & 1.95  & 1.78 & 0.47          & 0.46          & 1.88 & 1.85 & 1.90 & 2.03 & 0.45 & 0.44 \\
SinDDM      & 2.58 & 2.29 & 2.52  & 2.24 & 0.46          & 0.46          & 2.72 & 2.23 & 3.20 & 2.48 & 0.44 & 0.42 \\
\midrule
OUR-GAN     & \textbf{1.36} & \textbf{1.48} & \textbf{0.94} & \textbf{1.26} & 0.43 & 0.41 & \textbf{1.22} & \textbf{1.50} & \textbf{0.97} & \textbf{1.48} & 0.40 & 0.38\\
\bottomrule
\end{tabular}
\end{sc}
\end{small}
\end{center}
\vskip -0.1in
\caption{
Quantitative results of 4K non-repetitive image synthesis.
We measured SIFID at 4K resolution according to the method calculated in \cref{tab:4K_image_synthesis}, and LPIPS of all pairs of 50 synthesized images at 4K resolution.
↓ indicates that the lower the better while ↑ indicates the higher the better.}
\label{tab:Comparison_global}
\end{table*}

\begin{figure*}[!htb]
\vskip 0.1in
\begin{center}
\centerline{\includegraphics[width=0.9\textwidth]{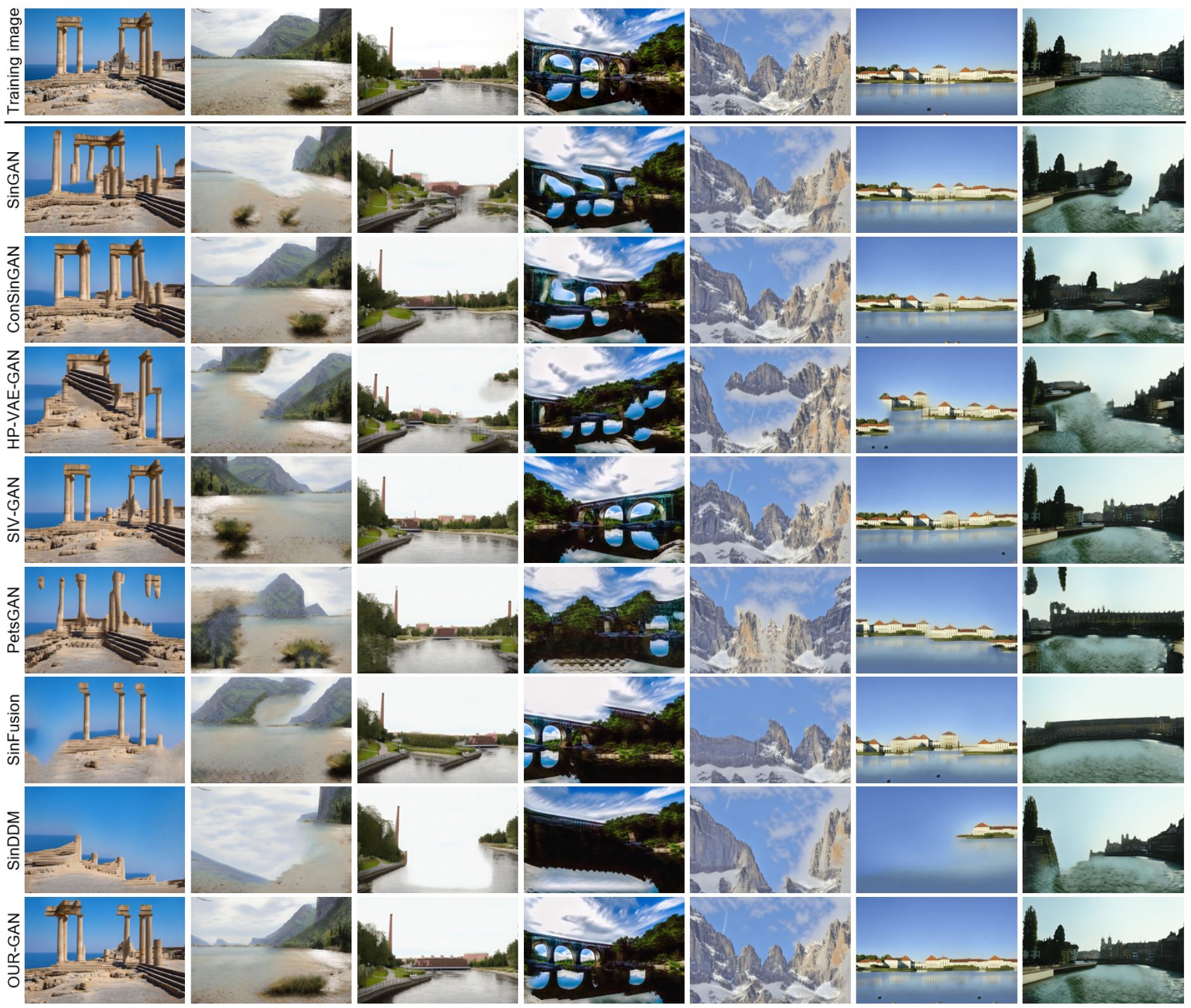}}
\caption{ 4K images synthesized from various training examples by OUR-GAN and seven baseline models. ConSinGAN and SIV-GAN exhibited limited diversity, often producing images with little difference from the training image or its horizontal flip. HP-VAE-GAN exhibited high diversity but poor visual coherence. PetsGAN, SinFusion, and SinDDM produced images with blended textures and unnatural layouts. In contrast, OUR-GAN consistently produces results with global coherence, regardless of the training image.
}
\label{fig:Comparison_global_2}
\end{center}
\vskip -0.2in
\end{figure*}


To evaluate the memory efficiency of OUR-GAN, we compared its GPU memory requirement with those of the baseline models. For empirical analysis, we measured the maximum memory consumption of OUR-GAN and the baseline models using their open-source implementations \cite{InGAN-src, SinGAN-src}. For theoretical analysis, we estimated the memory requirement for training by counting the parameters, feature values, and gradients of the entire network. To estimate the amount of memory needed for synthesis, we counted the parameters of the entire network and the input and output feature values of the largest layer since the memory for feature maps can be reused. The measured memory consumption is significantly higher than the theoretical estimate because the PyTorch framework and CUDA library require additional memory for efficient computation.

\cref{tab:memory_efficiency} presents the memory consumption of the one-shot synthesis models for training and synthesis. The baseline models used 21.32 $\sim$ 22.46 GB of GPU memory at maximum for training and synthesis. On the other hand, OUR-GAN used 5.59 GB for training and 4.29 GB for synthesis at maximum, which are remarkably lower than those of the baseline models. OUR-GAN also exhibited significantly lower memory requirements than the baseline models in theoretical analysis.

\begin{figure}[t] 
\begin{center}
\centerline{\includegraphics[width=\columnwidth]{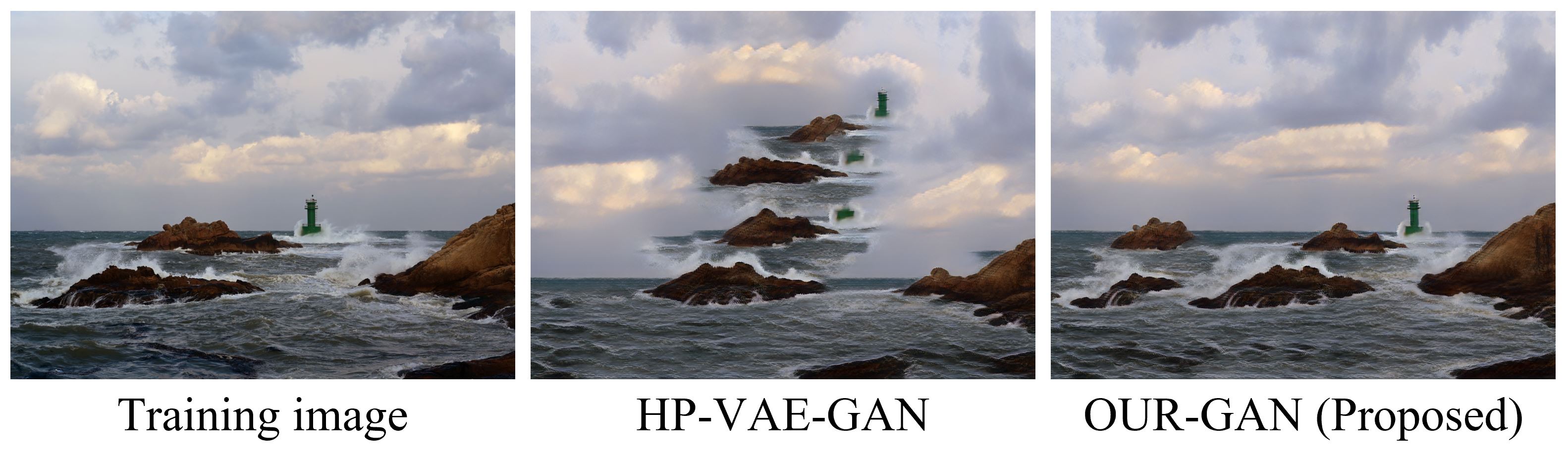}}
\caption{The effect of vertical coordinate convolution. OUR-GAN exhibits improved visual coherence than HP-VAE-GAN by exploiting spatial bias with vertical coordinate convolution.}
\label{fig:VCoord_coherence}
\end{center}
\end{figure}

We quantitatively compared OUR-GAN with the baseline models. \cref{tab:4K_image_synthesis} represents the SIFID of the one-shot synthesis models measured from the feature maps of the 5th and 13th convolution layers of the Inception network. For simplicity, we count an Inception block as a single layer. OUR-GAN exhibited significantly lower SIFID than the baseline models in all configurations. These results indicate that OUR-GAN produces images whose properties are closer to the training image than the baseline models.
\begin{figure}[!ht] 
\begin{center}
\centerline{\includegraphics[width=0.85\columnwidth]{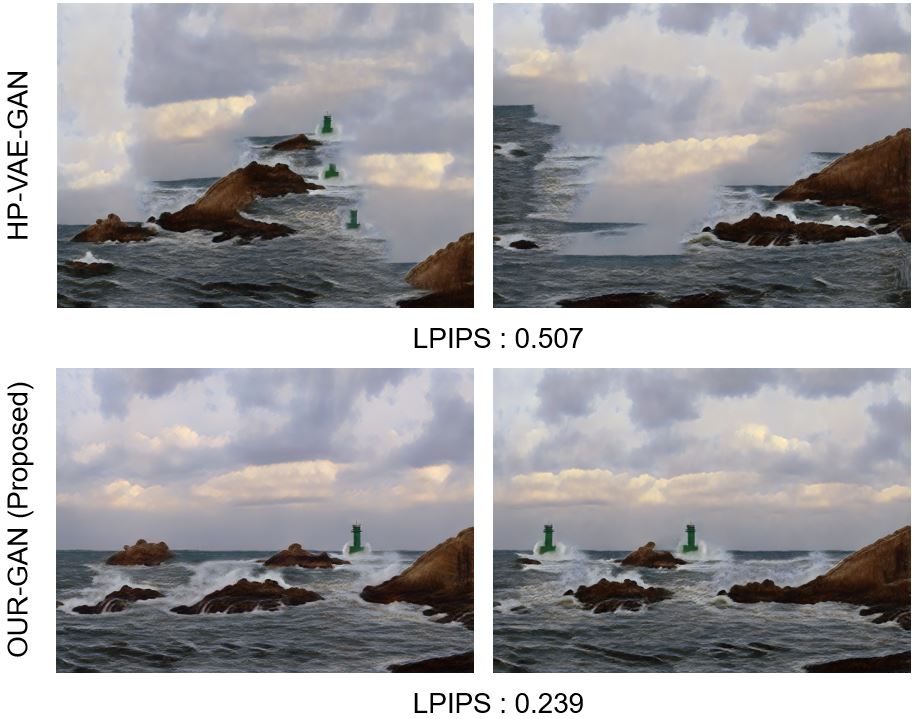}}
\caption{A limitation of LPIPS as a diversity metric. Visually incoherent images achieve higher LPIPS.}
\label{fig:LPIPS_bad}
\end{center}
\end{figure}
\cref{fig:4K_scenery_synthesis} displays the training image and the examples of the 4K images synthesized by OUR-GAN and the baseline models. InGAN \cite{InGAN} failed to synthesize a visually plausible UHR image containing large shapes because it synthesizes images by combining patches learned from relatively small training images.
SinGAN successfully generated large patterns but did not catch the fine details of the structure.
However, OUR-GAN successfully synthesized high-quality images with visually coherent shapes and fine details.
More qualitative comparison results are presented in \cref{appendix:Comparison_4K}.

\begin{figure*}[!htb]
\begin{center}
\centerline{\includegraphics[width=0.85\textwidth]{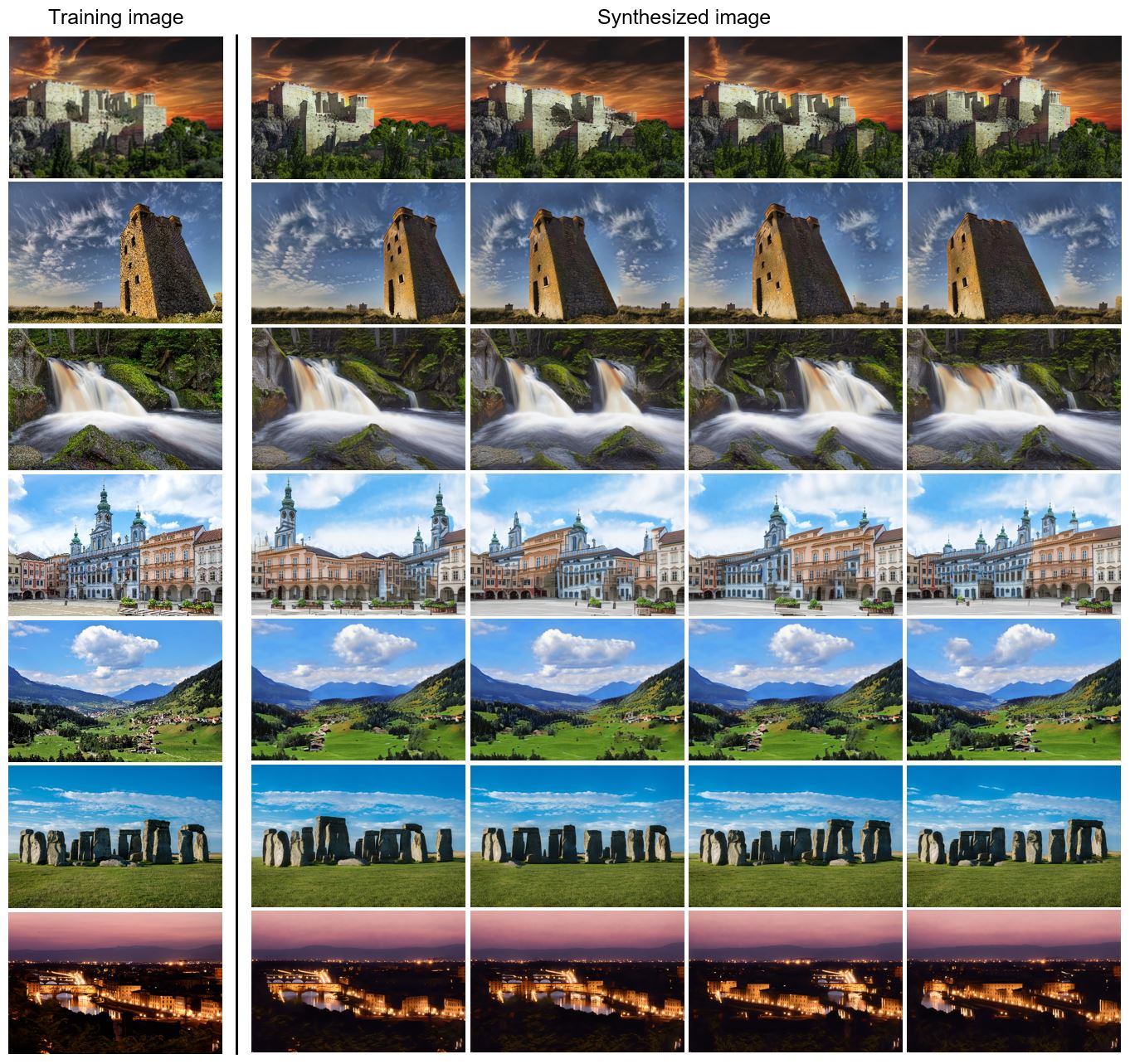}}
\caption{Multiple 4K images synthesized by OUR-GAN. OUR-GAN synthesizes high-quality images with diverse layouts. In the sixth row, the shape, size, number, and location of Stonehenge were varied, while the color and texture remained the same.}
\label{fig:4K_diversity}
\end{center}
\end{figure*}

\begin{figure*}[!ht]
\begin{center}
\centerline{\includegraphics[width=0.95\textwidth]{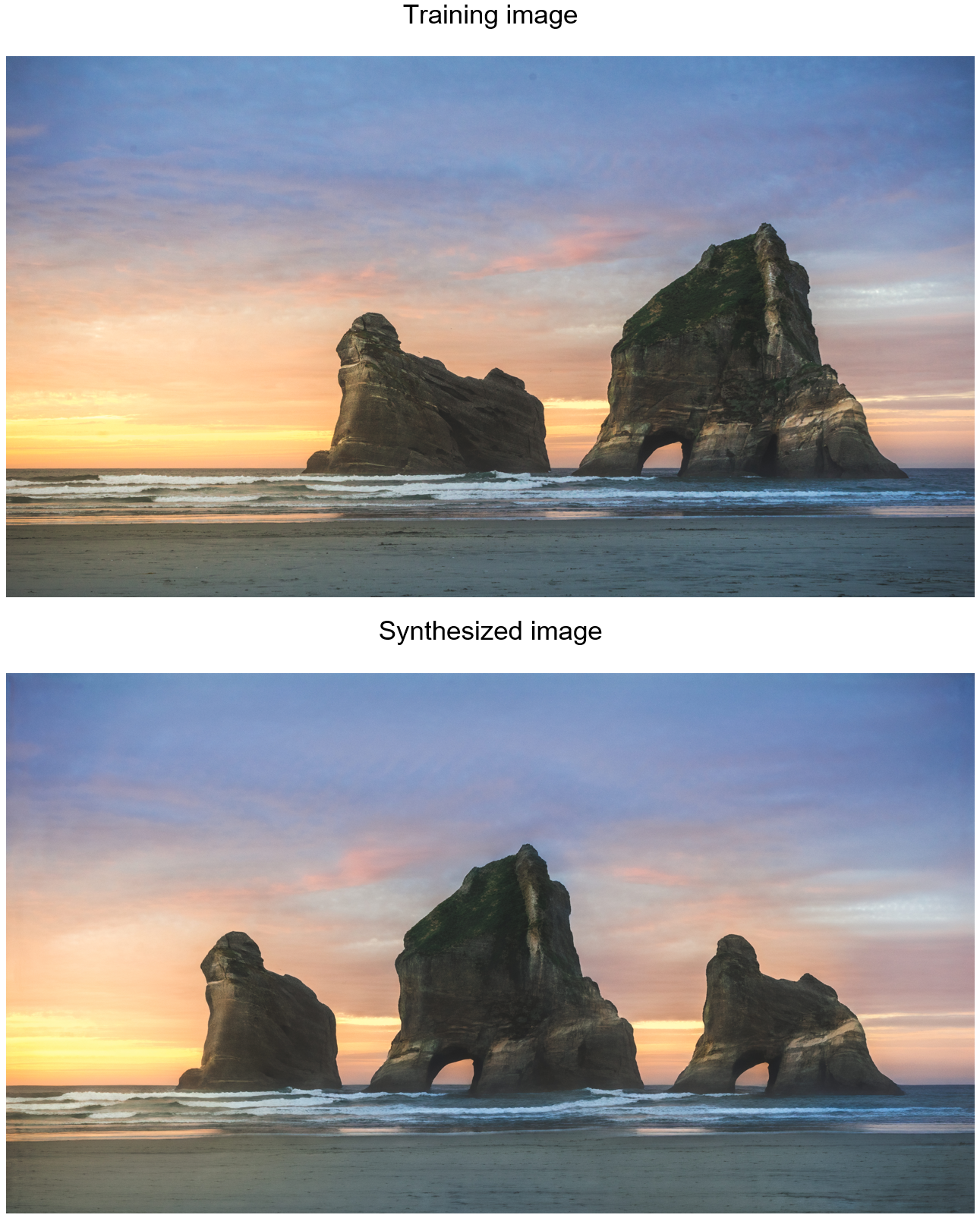}}
\caption{A 16K (16,384x9,152) UHR image synthesized by OUR-GAN-16K. The size of the biggest island at the center of the synthesized image is $4,913\times4,314$. OUR-GAN-16K can synthesize not only large-sized images but also images containing large shapes. }
\label{fig:16K_Image_Synthesis_2}
\end{center}
\end{figure*}

\subsection{Visual coherence and diversity} \label{result_vcoord}

We evaluated the image quality and diversity of OUR-GAN at 4K resolution. OUR-GAN combines a global structure generator and multiple super-resolution modules. The global structure generator is composed of HP-VAE-GAN extended by applying vertical coordinate convolution as described in \cref{subsection:global_structure_generation}. We compared the global structure generator of OUR-GAN with seven one-shot synthesis models: SinGAN \cite{SinGAN}, ConSinGAN \cite{ConSinGAN}, SIV-GAN \cite{SIV-GAN}, PetsGAN \cite{PetsGAN}, SinFusion \cite{SinFusion}, SinDDM \cite{SinDDM} and vanilla HP-VAE-GAN \cite{HP-VAE-GAN} that does not apply vertical coordinate convolution.
We excluded InGAN \cite{InGAN} as it is a conditional model that remaps the input image to the one target output image, not from noise.
For the super-resolution models, we implemented two sets of in-memory and subregion-wise super-resolution modules based on ESRGAN \cite{ESRGAN} and SwinIR \cite{SwinIR}, respectively.
We evaluated the image quality and diversity of all possible combinations of the one-shot synthesis models and the super-resolution modules  in SIFID and LPIPS.

\par

\cref{tab:Comparison_global} presents the result of quantitative evaluation on the ST4K and RAISE datasets. 
In regard to image quality, the global structure generator of OUR-GAN exhibited the lowest SIFID in all configurations.
Vanilla HP-VAE-GAN exhibited the highest LPIPS, indicating the highest diversity. However, vanilla HP-VAE-GAN did not reliably synthesize visually plausible images and often produced visually incoherent images as shown in Figures \ref{fig:Comparison_global}, \ref{fig:Comparison_global_2} and \ref{fig:VCoord_coherence}. As shown in \cref{fig:LPIPS_bad}, LPIPS can also be measured high for visually incoherent images. We measured the LPIPS between the two synthesized images. HP-VAE-GAN synthesizes visually incoherent patterns, but the resulting images have a high LPIPS. A high LPIPS does not guarantee the desired diversity since it does not consider the visual coherence of patterns.
The LPIPS of OUR-GAN is comparable to those of the other models. These results suggest that the proposed vertical coordinate convolution improves image quality with minimal loss of diversity. Regarding the super-resolution models, SwinIR exhibited higher image quality and diversity than ESRGAN in all configurations.

\cref{fig:Comparison_global} and \cref{fig:Comparison_global_2} present multiple 4K images synthesized by OUR-GAN and the baseline models. Because the baseline models have limited output resolutions, we increased their output resolutions by combining them with the second and third modules of OUR-GAN to produce 4K images. In \cref{fig:Comparison_global}, SinGAN \cite{SinGAN} failed to generate visually plausible images. ConSinGAN \cite{ConSinGAN}, an extension of SinGAN with improved training techniques, exhibits limited diversity. HP-VAE-GAN \cite{HP-VAE-GAN} exhibited higher diversity but poor visual coherence.
In \cref{fig:Comparison_global_2}, PetsGAN \cite{PetsGAN}, SinFusion \cite{SinFusion}, and SinDDM \cite{SinDDM} synthesize images with mixed textures and unnatural layouts. For example, in the second column, the mountains and sand are floating in the sky, and the boundaries between them are unclear, resulting in a mixture of mountains, sand, and sky.
Additionally, as acknowledged in their paper as a limitation, SinDDM produces images in which objects are often omitted.
In contrast, OUR-GAN synthesizes images with reasonable layouts and diversity by successfully changing the shape and location of pillars, bridges, mountains, etc.


\cref{fig:4K_diversity} presents multiple 4K images synthesized by OUR-GAN, demonstrating its diversity and visual coherence. 
These images have diverse and visually plausible layouts while preserving the semantic content and style of the training image, such as class identity, color, and texture. For example, the synthesized images in the first row show castles with different shapes, sizes, and locations from other images while preserving the visual appearance of the castle in the training image, resulting in a visually plausible image.

As mentioned earlier, we also qualitatively compared OUR-GAN with vanilla HP-VAE-GAN to test the effect of the proposed vertical coordinate convolution. As shown in \cref{fig:VCoord_coherence}, applying vertical coordinate convolution significantly improved the visual coherence of scenery images by assisting the model to learn visual elements associated with their vertical coordinates. 


\subsection{One-shot 16K image synthesis}
\label{result_16K}
We also synthesized 16K scenery images with OUR-GAN composed of a global structure generator, an in-memory super-resolution module, and two subregion-wise super-resolution modules. We call this model OUR-GAN-16K. As ST4K and RAISE are composed of 4K images, we collected 16K scenery images from the Internet to train OUR-GAN-16K. \cref{fig:16K_Image_Synthesis_2} presents a 16K synthesized image and its training image. The size of the biggest island at the center of the image is approximately $4,913\times4,314$. OUR-GAN-16K successfully synthesized non-repetitive high-fidelity 16K images maintaining both visual coherence and fine details. OUR-GAN-16K used 19.6 GB for training and 12.5 GB for synthesis. Our demo page presents other 16K synthesized samples.

\section{Conclusion}

In this paper, we proposed a One-shot Ultra-high-Resolution GAN (OUR-GAN) framework that synthesizes high-fidelity non-repetitive UHR images with a resolution of 4K $\sim$ 16K and is trainable from a single image. OUR-GAN generates diverse and globally coherent large shapes with fine details and maintains long-range coherence.
To synthesize UHR images with limited GPU memory, OUR-GAN first created a diverse and visually coherent initial image and gradually increased the resolution through in-memory and subregion-wise super-resolution modules based on SwinIR. To improve visual coherence, we combined vertical coordinate convolution with HP-VAE-GAN. We also minimized the memory overhead by estimating the overlap of the subregion-wise super-resolution using the effective receptive field of the super-resolution model. The experimental results confirm the effectiveness of the proposed framework in generating high-quality UHR images with 4K $\sim$ 16K resolutions and improving image quality and visual coherence.

\bibliographystyle{elsarticle-num} 
\bibliography{reference}

\newpage
\appendix
\numberwithin{algorithm}{section}
\numberwithin{equation}{section}
\numberwithin{figure}{section}
\numberwithin{table}{section}
\onecolumn

\section{Implementation of the baseline models}\label{appendix:baselines}
We implemented the baseline models with open sources.

These are the links to the open sources we employed.
\begin{itemize}
    \item SinGAN: https://github.com/tamarott/SinGAN
    \item ConSinGAN: https://github.com/tohinz/ConSinGAN
    \item HP-VAE-GAN: https://github.com/shirgur/hp-vae-gan
    \item InGAN: https://github.com/Caenorst/InGAN/tree/py3
    \item PetsGAN: https://github.com/zhangzc21/petsgan
    \item SinFusion: https://github.com/yanivnik/sinfusion-code
    \item SinDDM: https://github.com/fallenshock/SinDDM
\end{itemize}
Note that we implemented SIV-GAN with codes from the author, which have not been released.

\clearpage

\section{ST4K dataset}\label{appendix:ST4K}
\begin{figure}[htb] 
\vskip 0.2in
\begin{center}
\centerline{\includegraphics[width=\textwidth]{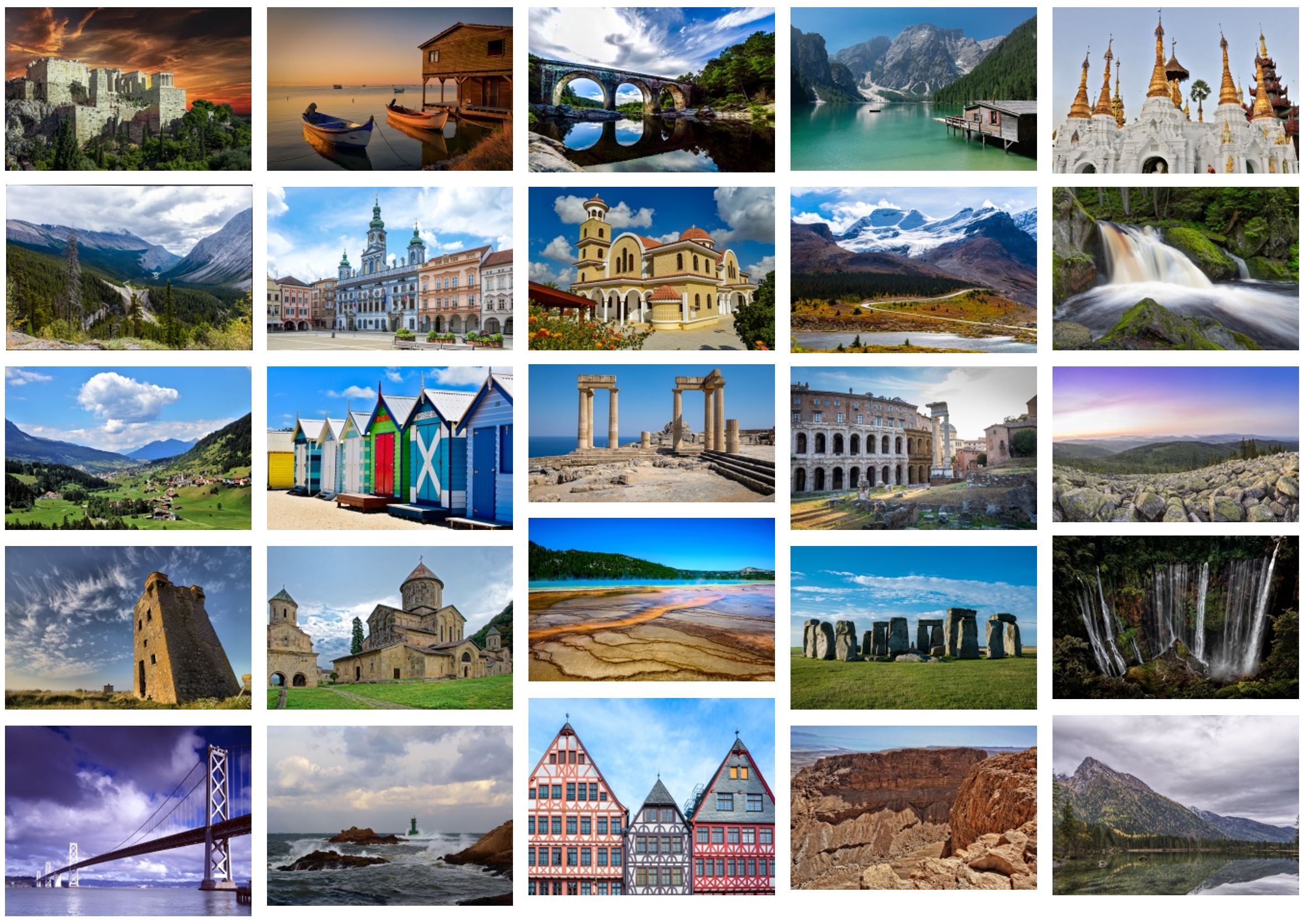}}
\caption{Scenery images of ST4K.}
\label{fig:ST4K-Scenery}
\end{center}
\vskip -0.2in
\end{figure}

\begin{figure}[htb] 
\vskip 0.2in
\begin{center}
\centerline{\includegraphics[width=\textwidth]{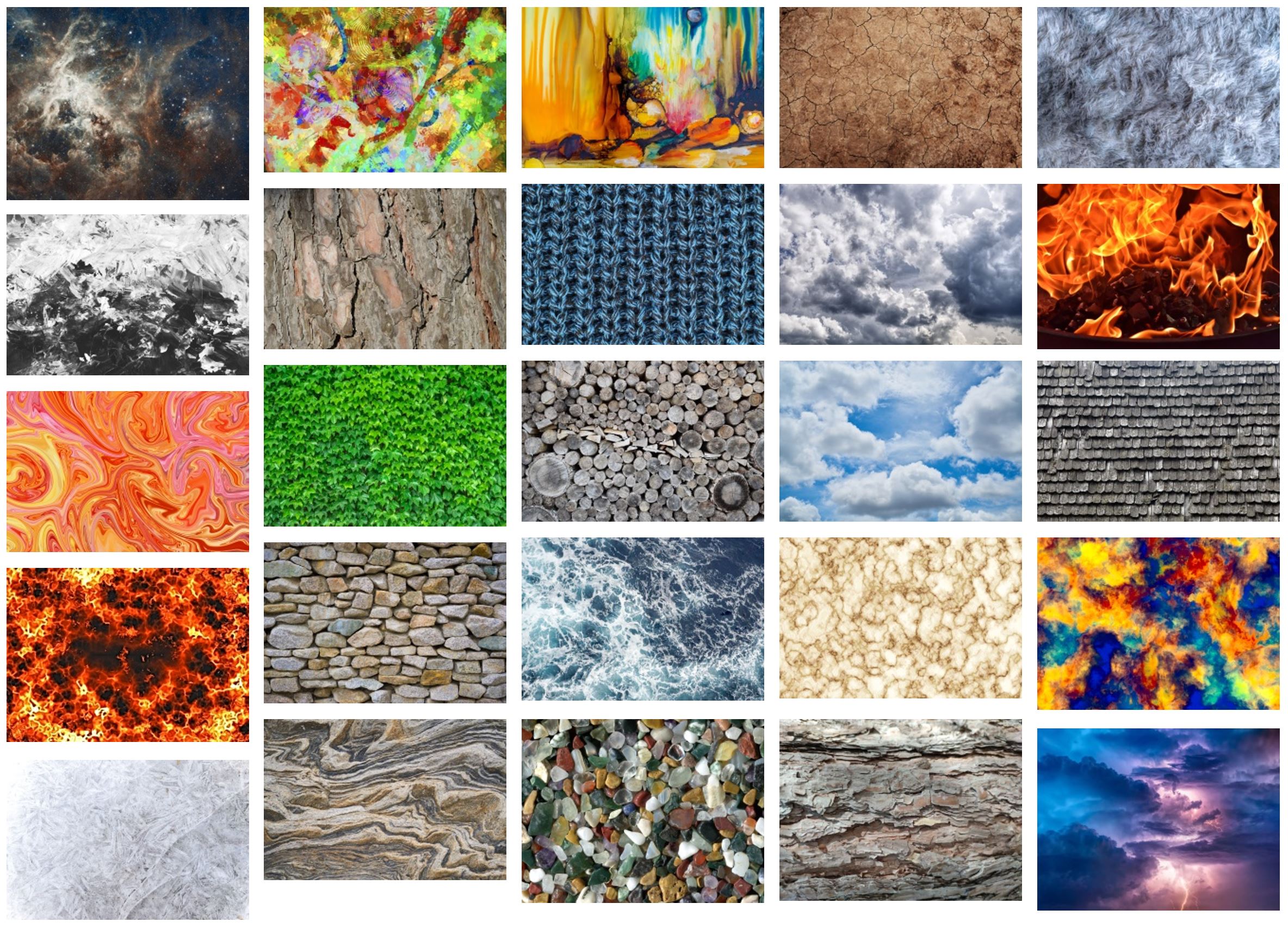}}
\caption{Texture images of ST4K.}
\label{fig:ST4K-Texture}
\end{center}
\vskip -0.2in
\end{figure}

We collected 4K or higher images from Pixabay\footnote{https://pixabay.com/}.
About scenery and texture categories, we tried to gather diverse images.
To equalize the image size, we resized the images larger than 4K resolution to 4K.
\cref{fig:ST4K-Scenery} and \cref{fig:ST4K-Texture} shows all images of ST4K.

\clearpage

\section{Images selected from RAISE}\label{appendix:RAISE}

\begin{figure}[h!] 
\vskip 0.2in
\begin{center}
\centerline{\includegraphics[width=\textwidth]{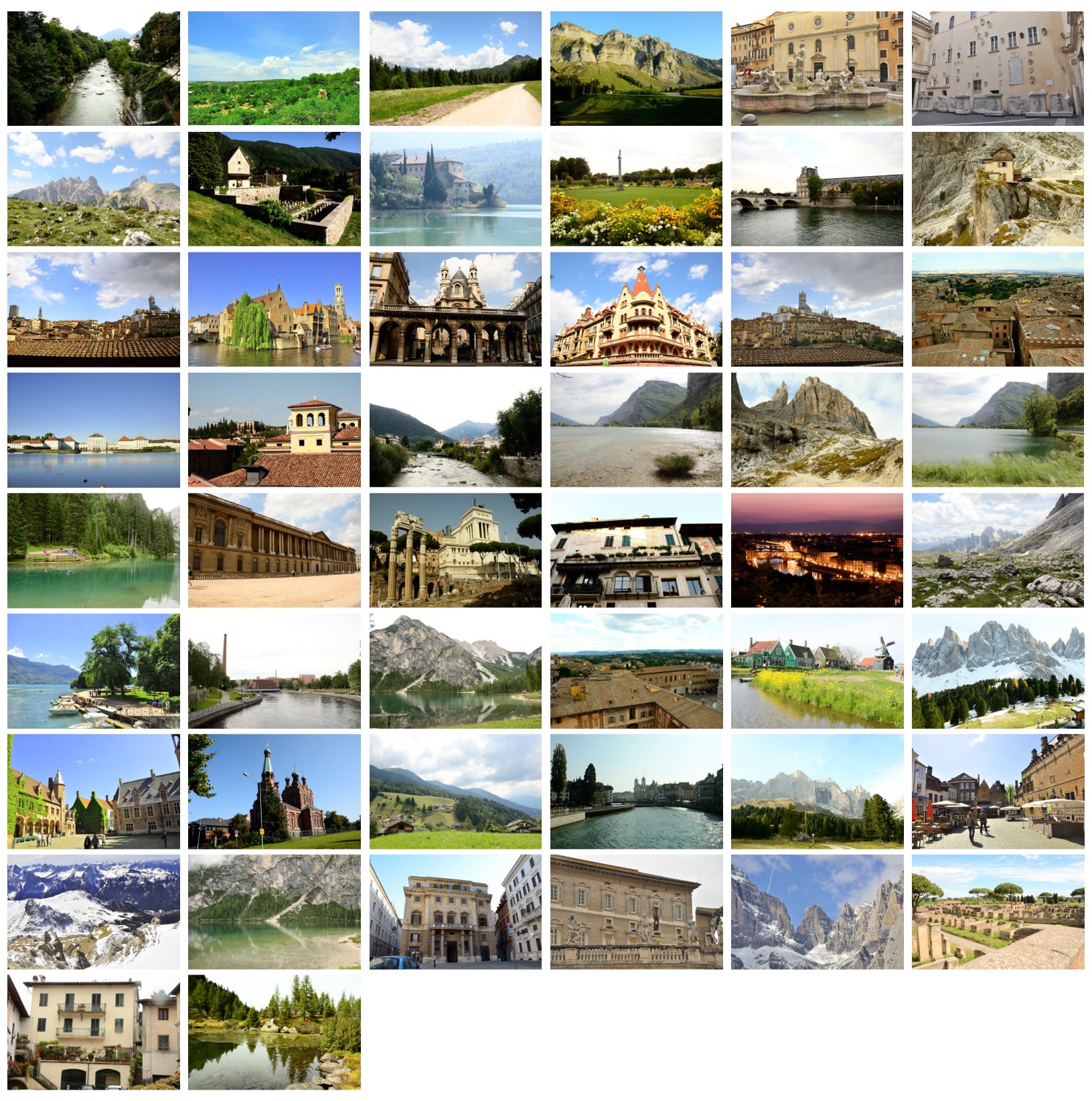}}
\caption{50 images selected from RAISE}
\label{fig:RAISE}
\end{center}
\vskip -0.2in
\end{figure}

\clearpage


\clearpage

\section{More qualitative results}\label{appendix:samples}

\begin{figure}[h] 
\begin{center}
\centerline{\includegraphics[width=0.73\textwidth]{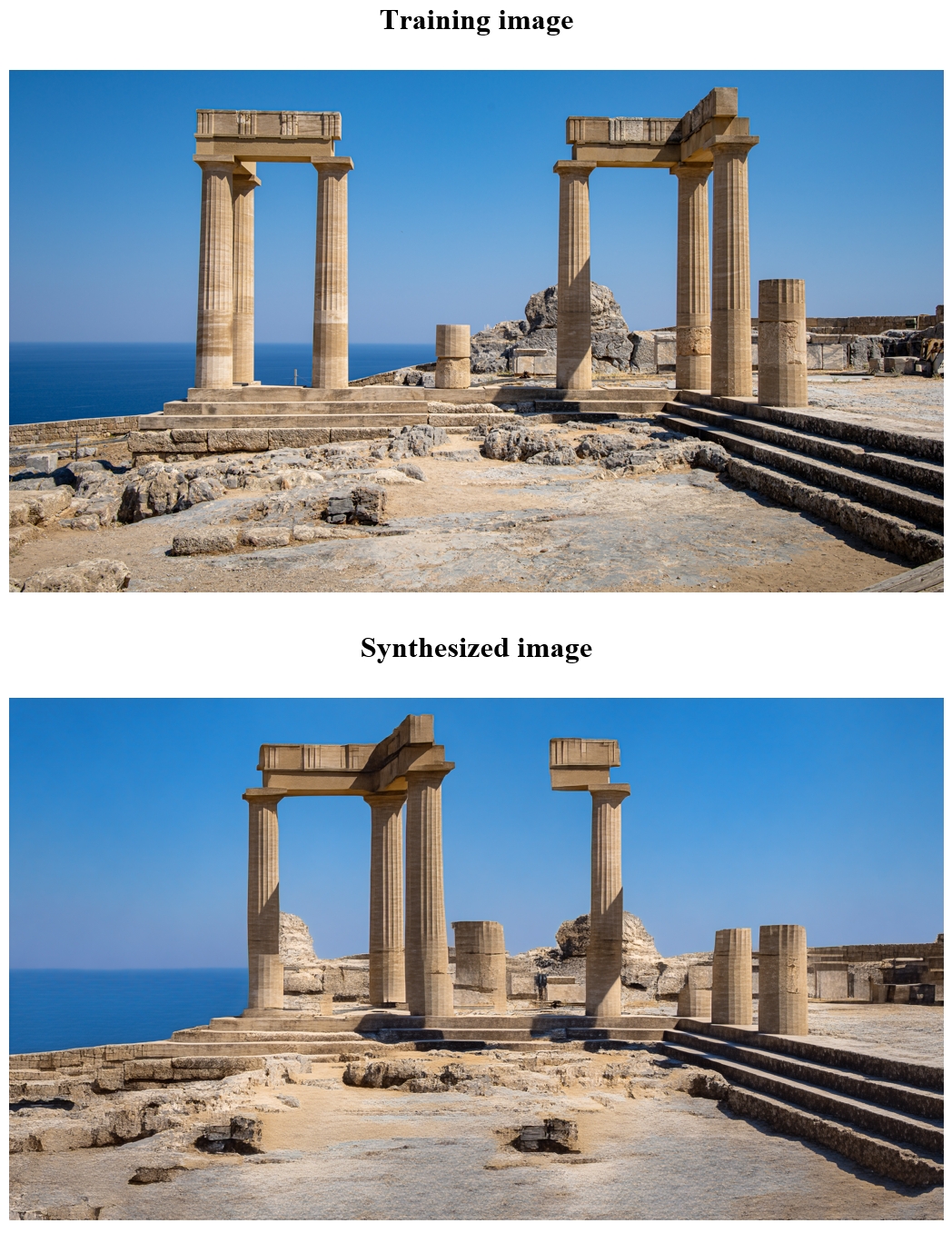}}
\caption{4K image synthesized by OUR-GAN.}
\label{fig:4K_samples_1}
\end{center}
\end{figure}

\begin{figure}[h] 
\begin{center}
\centerline{\includegraphics[width=0.9\textwidth]{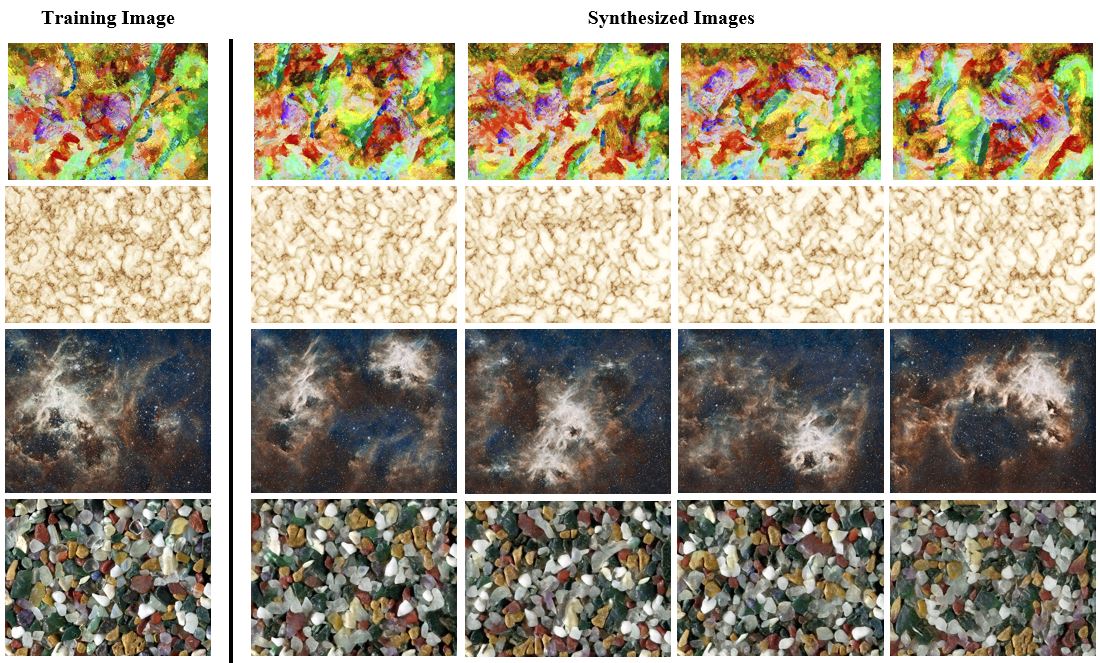}}
\caption{Texture images synthesized by OUR-GAN}
\label{fig:random_samples_3}
\end{center}
\end{figure}

\clearpage

\section{More qualitative comparison results}\label{appendix:Comparison_4K}

\begin{figure}[h] 
\begin{center}
\centerline{\includegraphics[width=0.9\textwidth]{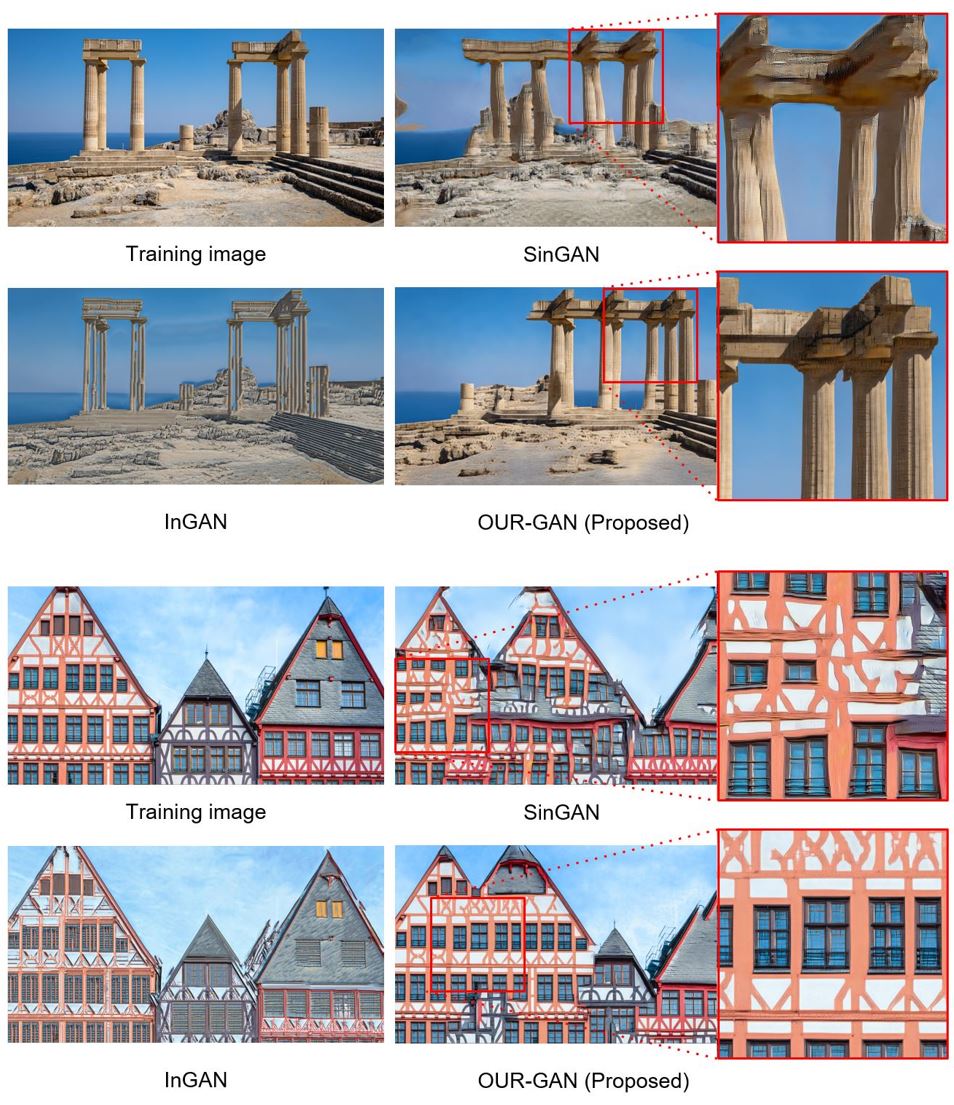}}
\caption{Additional qualitative comparison results of non-repetitive 4K Image Synthesis}
\label{fig:FigureE.1_Comparison_4K_2}
\end{center}
\vskip -0.9in
\end{figure}

\clearpage

\section{Training}\label{appendix:training}
\subsection{Training in the first step}
\begin{figure}[h] 
\begin{center}
\centerline{\includegraphics[width=\textwidth]{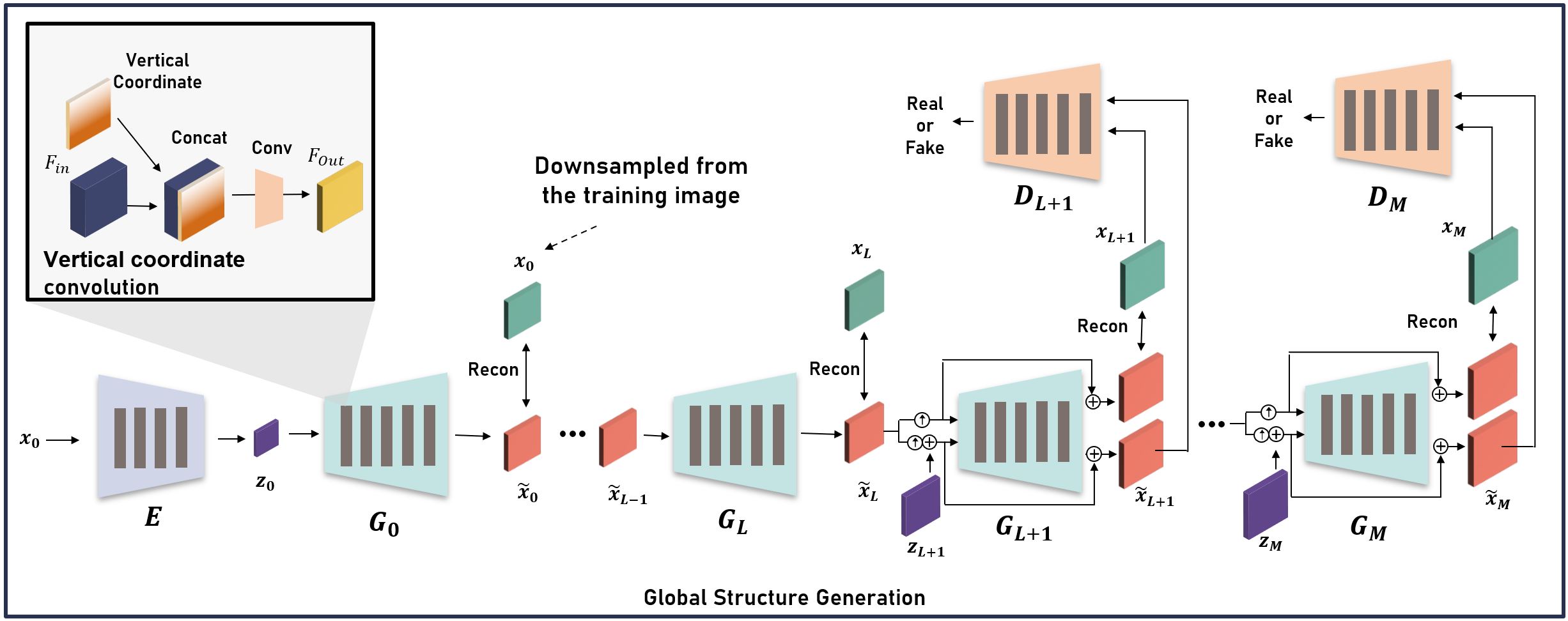}}
\caption{Training procedure in the 1st step.}
\label{fig:train_1st}
\end{center}
\end{figure}

We present the training procedure and algorithm in \cref{fig:train_1st} and \cref{alg:train_1st}.
We compute the reconstruction loss $\mathcal{L}_{recon}$ with MSE, the adversarial loss $\mathcal{L}_{adv}$ with WGAN-GP loss, and the KL loss $\mathcal{L}_{\mathbb{KL}}$ following \cite{PatchVAE}.

\hfill

\begin{algorithm}[h]
    \caption{Training algorithm in the 1st step.}
    \label{alg:train_1st}
\begin{algorithmic}
    \STATE {\bfseries Single training image at scale m:} $x_m$
    \STATE {\bfseries Total number of scales:} $M$
    \STATE {\bfseries The last scale applying patchVAE:} $L$
    \STATE {\bfseries Encoder:} $E$
    \STATE {\bfseries Generator at scale m:} $G_m$
    \STATE {\bfseries Discriminator at scale m:} $D_m$
    \STATE {\bfseries Random noise at scale m:} $z_m$

    \hfill
    
    \FOR{$m=0$ {\bfseries to} $M$}
    \FOR{$i=0$ {\bfseries to} number of epochs}


    \IF{$m = 0$}
    \STATE $f$ $\leftarrow$ {$E(x_0)$}
    \STATE Samples latent codes $z_0$ with $f$
    \STATE $\tilde{x}_0 \leftarrow G_0(z_0)$
    \ENDIF

    \IF{$1 \leq m \leq L$}
    \STATE $\tilde{x}_m \leftarrow \uparrow\tilde{x}_{m-1} + G_m(\uparrow\tilde{x}_{m-1})$
    \ENDIF

    \IF{$L < m \leq M$}
    \STATE $\tilde{x}_m \leftarrow \uparrow \tilde{x}_{m-1} + G_m(\uparrow\tilde{x}_{m-1} + z_n)$
    \STATE $real/fake \leftarrow D_m(\tilde{x}_{m}), D_m(x_{m})$
    \ENDIF
    
    \hfill


    \IF{$m = 0$}
    \STATE $\mathcal{L} \leftarrow \mathcal{L}_{recon}(\tilde{x}_0, x_0) + \beta_{vae} \mathcal{L}_{\mathbb{KL}}(x_0)$
    \STATE Backpropagate $\mathcal{L}$
    \STATE Update $E$ and $G_0$
    \ENDIF

    \IF{$1 \leq m \leq L$}
    \STATE $\mathcal{L} \leftarrow \mathcal{L}_{recon}(\tilde{x}_n, x_n) + \mathcal{L}_{recon}(\tilde{x}_0, x_0) + \beta_{vae} \mathcal{L}_{\mathbb{KL}}(x_0)$
    \STATE Backpropagate $\mathcal{L}$
    \STATE Update $E$, $G_0$, and $G_m$ 
    \ENDIF

    \IF{$L < m \leq M$}
    \STATE $\mathcal{L} \leftarrow \mathcal{L}_{recon}(\tilde{x}_n, x_n) + \beta_{adv} \mathcal{L}_{adv}(z, x_n)$
    \STATE Backpropagate $\mathcal{L}$
    \STATE Update $G_m$ and $D_m$
    \ENDIF
    \ENDFOR
    \ENDFOR

\end{algorithmic}
\end{algorithm}

\clearpage

\subsection{Training in the second and third steps}

\begin{figure}[h] 
\begin{center}
\centerline{\includegraphics[width=0.7\textwidth]{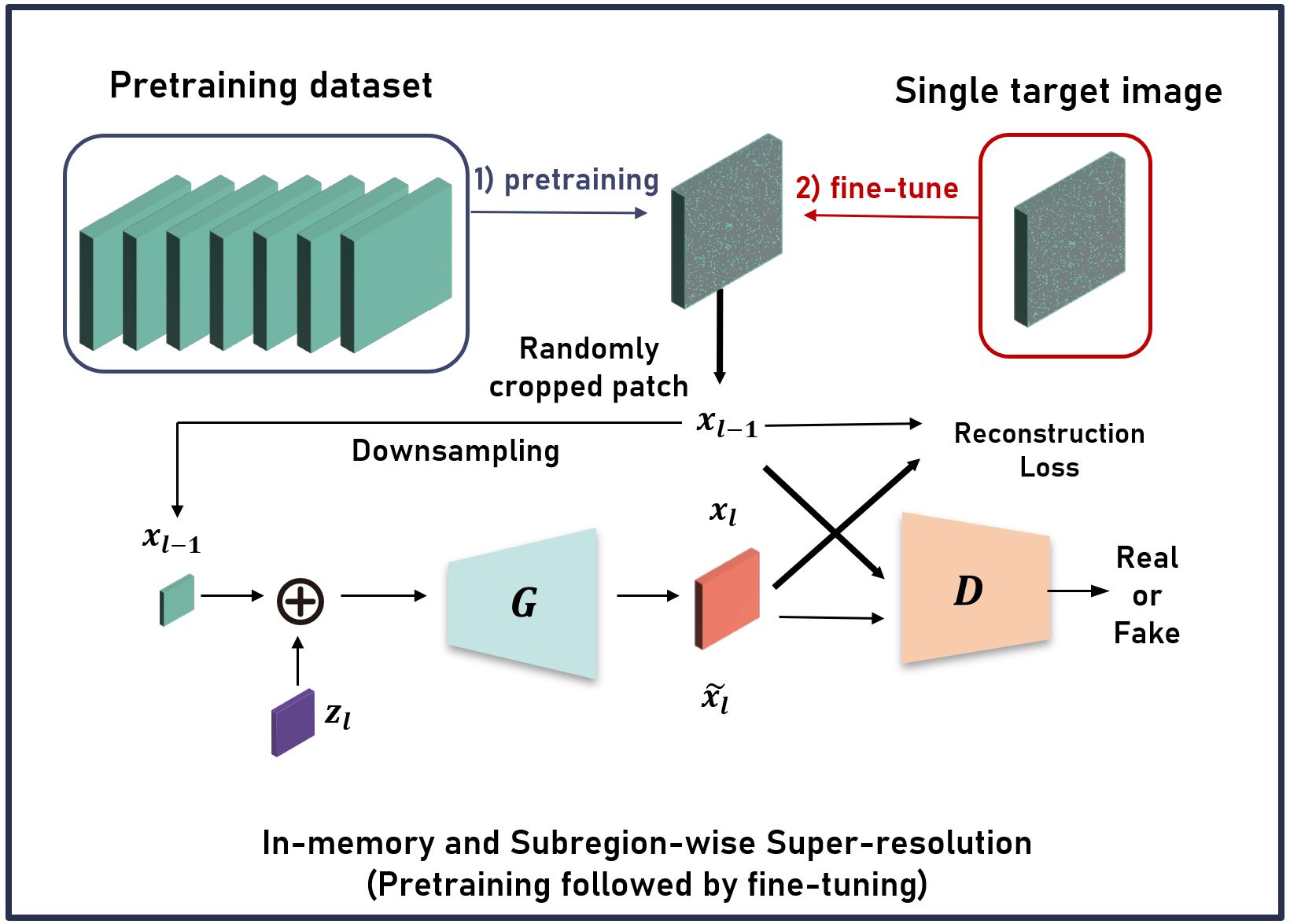}}
\caption{Training procedure in the 2nd and 3rd steps.}
\label{fig:train_2&3_st}
\end{center}
\end{figure}

We present the training procedure and algorithm in \cref{fig:train_2&3_st} and \cref{alg:ESRGAN-pre-train}, \cref{alg:ESRGAN-fine-tune}, \cref{alg:SwinIR-pre-train},  \cref{alg:SwinIR-fine-tune}. 
We first pre-train the super-resolution model(generator) with a large-scale dataset.
Then, we fine-tune the super-resolution model(generator) with a single target image.
ESRGAN and SwinIR are optimized by a reconstruction loss ($\mathcal{L}_{recon}$), perceptual loss ($\mathcal{L}_{perceptual}$), and adversarial loss. The Reconstruction loss and perceptual loss are computed with L1 distance. For the adversarial loss, ESRGAN is trained with RaGAN loss and SwinIR is trained with vanilla GAN loss. We trained SwinIR by using the same degradation model as BSRGAN \cite{BSRGAN}.


\hfill 

\clearpage

\begin{algorithm}[h]
    \caption{ESRGAN pre-training algorithm in the 2nd and 3rd steps.}
    \label{alg:ESRGAN-pre-train}
\begin{algorithmic}
    \STATE {\bfseries Dataset:} $X = \{x^1, x^2, ..., x^n\}$
    \STATE {\bfseries Generator:} $G$
    \STATE {\bfseries Discriminator:} $D$
    \STATE {\bfseries Feature extractor:} $F$
    \STATE {\bfseries Augmentation:} $T$
    \STATE {\bfseries Downsampling:} $S$

    \hfill
    
    \FOR{$i=0$ {\bfseries to} number of steps}
    \STATE Samples a mini-batch $x_{hr}$ from $X$

    \hfill

    \STATE $x_{lr} \leftarrow S(T(x_{hr}))$
    \STATE $x_{sr} \leftarrow G(x_{lr})$

    \hfill
    
    \STATE $\mathcal{L}_{G} \leftarrow \mathcal{L}_{recon}(x_{lr}, x_{sr})$
    \STATE Backpropagate $\mathcal{L}_{G}$
    \STATE Update $G$
    \ENDFOR

    \hfill

    \FOR{$j=0$ {\bfseries to} number of steps}
    \STATE Samples a mini-batch $x_{hr}$ from $X$

    \hfill

    \STATE $x_{lr} \leftarrow S(T(x_{hr}))$
    \STATE $x_{sr} \leftarrow G(x_{lr})$
    \STATE $o_{sr}, o_{hr} \leftarrow D(x_{sr}), D(x_{hr})$
    \STATE $f_{sr}, f_{hr} \leftarrow F(x_{sr}), F(x_{hr})$
    
    \hfill
    
    \STATE $\mathcal{L}_{G} \leftarrow \mathcal{L}_{perceptual}(f_{lr}, f_{sr}) + \lambda \mathcal{L}_{G}^{Ra}(o_{sr}, o_{hr}) + \eta \mathcal{L}_{recon}(x_{lr}, x_{sr})$
    \STATE $\mathcal{L}_{D} \leftarrow \mathcal{L}_{D}^{Ra}(o_{sr}, o_{hr})$

    \hfill

    \STATE Backpropagate $\mathcal{L}_G$ and $\mathcal{L}_D$
    \STATE Update $G$ and $D$
    \ENDFOR
\end{algorithmic}
\end{algorithm}

\begin{algorithm}[h]
    \caption{ESRGAN fine-tuning algorithm in the 2nd and 3rd steps.}
    \label{alg:ESRGAN-fine-tune}
\begin{algorithmic}
    \STATE {\bfseries Dataset:} $X = \{x^1, x^2, ..., x^n\}$
    \STATE {\bfseries Generator:} $G$
    \STATE {\bfseries Discriminator:} $D$
    \STATE {\bfseries Feature extractor:} $F$
    \STATE {\bfseries Augmentation:} $T$
    \STATE {\bfseries Downsampling:} $S$

    \hfill
    
    \STATE {\bfseries Initialize G with pre-trained weights.}
    
    \hfill
    
    \FOR{$i=0$ {\bfseries to} number of steps}
    \STATE Samples a mini-batch $x_{hr}$ from $X$

    \hfill

    \STATE $x_{lr} \leftarrow S(T(x_{hr}))$
    \STATE $x_{sr} \leftarrow G(x_{lr})$
    \STATE $o_{sr}, o_{hr} \leftarrow D(x_{sr}), D(x_{hr})$
    \STATE $f_{sr}, f_{hr} \leftarrow F(x_{sr}), F(x_{hr})$
    
    \hfill
    
    \STATE $\mathcal{L}_{G} \leftarrow \mathcal{L}_{perceptual}(f_{lr}, f_{sr}) + \lambda \mathcal{L}_{G}^{Ra}(o_{sr}, o_{hr}) + \eta \mathcal{L}_{recon}(x_{lr}, x_{sr})$
    \STATE $\mathcal{L}_{D} \leftarrow \mathcal{L}_{D}^{Ra}(o_{sr}, o_{hr})$

    \hfill

    \STATE Backpropagate $\mathcal{L}_G$ and $\mathcal{L}_D$
    \STATE Update $G$ and $D$
    \ENDFOR
\end{algorithmic}
\end{algorithm}

\clearpage

\begin{algorithm}[h]
    \caption{SwinIR pre-training algorithm in the 2nd and 3rd steps.}
    \label{alg:SwinIR-pre-train}
\begin{algorithmic}
    \STATE {\bfseries Dataset:} $X = \{x^1, x^2, ..., x^n\}$
    \STATE {\bfseries Generator:} $G$
    \STATE {\bfseries Discriminator:} $D$
    \STATE {\bfseries Feature extractor:} $F$
    \STATE {\bfseries Degradation model:} $DM$

    \hfill
    
    \FOR{$i=0$ {\bfseries to} number of steps}
    \STATE Samples a mini-batch $x_{hr}$ from $X$

    \hfill

    \STATE $x_{lr} \leftarrow DM(x_{hr})$
    \STATE $x_{sr} \leftarrow G(x_{lr})$

    \hfill
    
    \STATE $\mathcal{L}_{G} \leftarrow \mathcal{L}_{recon}(x_{lr}, x_{sr})$
    \STATE Backpropagate $\mathcal{L}_{G}$
    \STATE Update $G$
    \ENDFOR

    \hfill

    \FOR{$j=0$ {\bfseries to} number of steps}
    \STATE Samples a mini-batch $x_{hr}$ from $X$

    \hfill

    \STATE $x_{lr} \leftarrow S(DM(x_{hr}))$
    \STATE $x_{sr} \leftarrow G(x_{lr})$
    \STATE $o_{sr}, o_{hr} \leftarrow D(x_{sr}), D(x_{hr})$
    \STATE $f_{sr}, f_{hr} \leftarrow F(x_{sr}), F(x_{hr})$
    
    \hfill
    
    \STATE $\mathcal{L}_{G} \leftarrow \mathcal{L}_{perceptual}(f_{lr}, f_{sr}) + \lambda \mathcal{L}_{G}(o_{sr}, o_{hr}) + \eta \mathcal{L}_{recon}(x_{lr}, x_{sr})$
    \STATE $\mathcal{L}_{D} \leftarrow \mathcal{L}_{D}^{Ra}(o_{sr}, o_{hr})$

    \hfill

    \STATE Backpropagate $\mathcal{L}_G$ and $\mathcal{L}_D$
    \STATE Update $G$ and $D$
    \ENDFOR
\end{algorithmic}
\end{algorithm}

\begin{algorithm}[h]
    \caption{SwinIR fine-tuning algorithm in the 2nd and 3rd steps.}
    \label{alg:SwinIR-fine-tune}
\begin{algorithmic}
    \STATE {\bfseries Dataset:} $X = \{x^1, x^2, ..., x^n\}$
    \STATE {\bfseries Generator:} $G$
    \STATE {\bfseries Discriminator:} $D$
    \STATE {\bfseries Feature extractor:} $F$
    \STATE {\bfseries Degradation model:} $DM$

    \hfill
    
    \STATE {\bfseries Initialize G with pre-trained weights.}
    
    \hfill
    
    \FOR{$i=0$ {\bfseries to} number of steps}
    \STATE Samples a mini-batch $x_{hr}$ from $X$

    \hfill

    \STATE $x_{lr} \leftarrow DM(x_{hr})$
    \STATE $x_{sr} \leftarrow G(x_{lr})$
    \STATE $o_{sr}, o_{hr} \leftarrow D(x_{sr}), D(x_{hr})$
    \STATE $f_{sr}, f_{hr} \leftarrow F(x_{sr}), F(x_{hr})$
    
    \hfill
    
    \STATE $\mathcal{L}_{G} \leftarrow \mathcal{L}_{perceptual}(f_{lr}, f_{sr}) + \lambda \mathcal{L}_{G}(o_{sr}, o_{hr}) + \eta \mathcal{L}_{recon}(x_{lr}, x_{sr})$
    \STATE $\mathcal{L}_{D} \leftarrow \mathcal{L}_{D}(o_{sr}, o_{hr})$

    \hfill

    \STATE Backpropagate $\mathcal{L}_G$ and $\mathcal{L}_D$
    \STATE Update $G$ and $D$
    \ENDFOR
\end{algorithmic}
\end{algorithm}

\clearpage

\section{The detailed structures of the generators and discriminators}\label{appendix:structure}
\begin{figure}[ht] 
\begin{center}
\centerline{\includegraphics[width=0.9\textwidth]{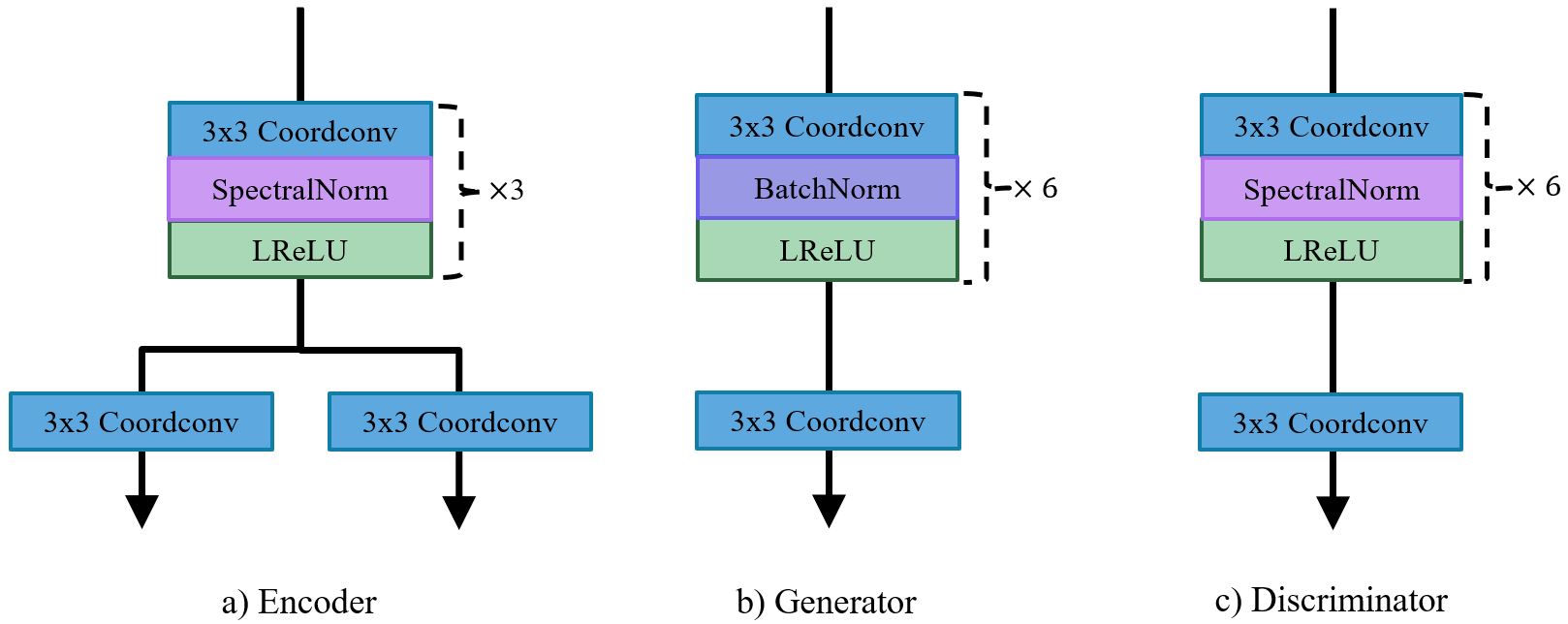}}
\caption{The detailed structure of the first step model}
\label{fig:First_step_Structure}
\end{center}
\end{figure}

\begin{figure}[hb!] 
\begin{center}
\centerline{\includegraphics[width=0.9\textwidth]{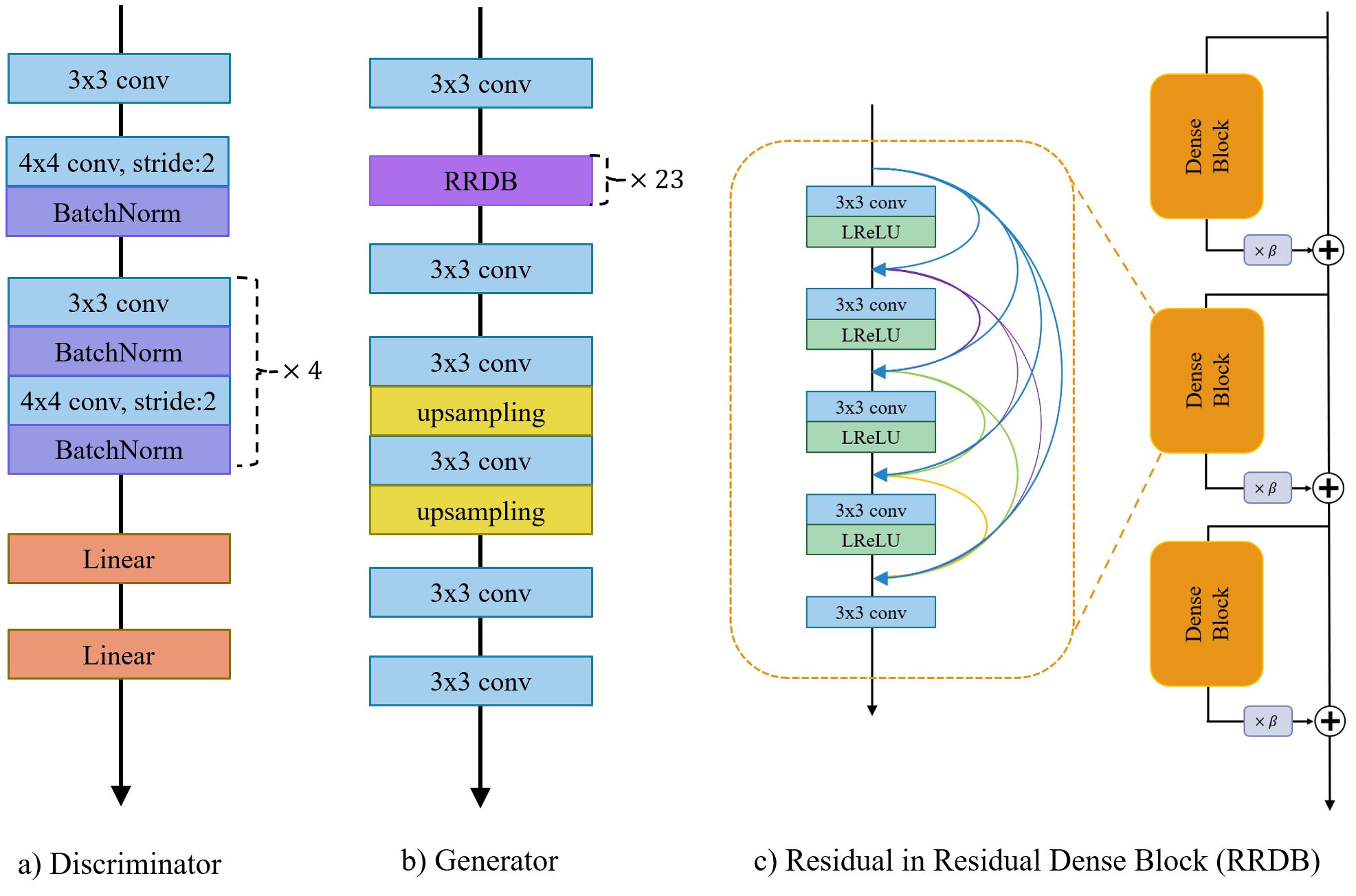}}
\caption{The detailed structure of the ESRGAN. 
We omitted activation functions for simplicity.}
\label{fig:ESRGAN_Structure}
\end{center}
\end{figure}

\begin{figure}[h] 
\begin{center}
\centerline{\includegraphics[width=0.9\textwidth]{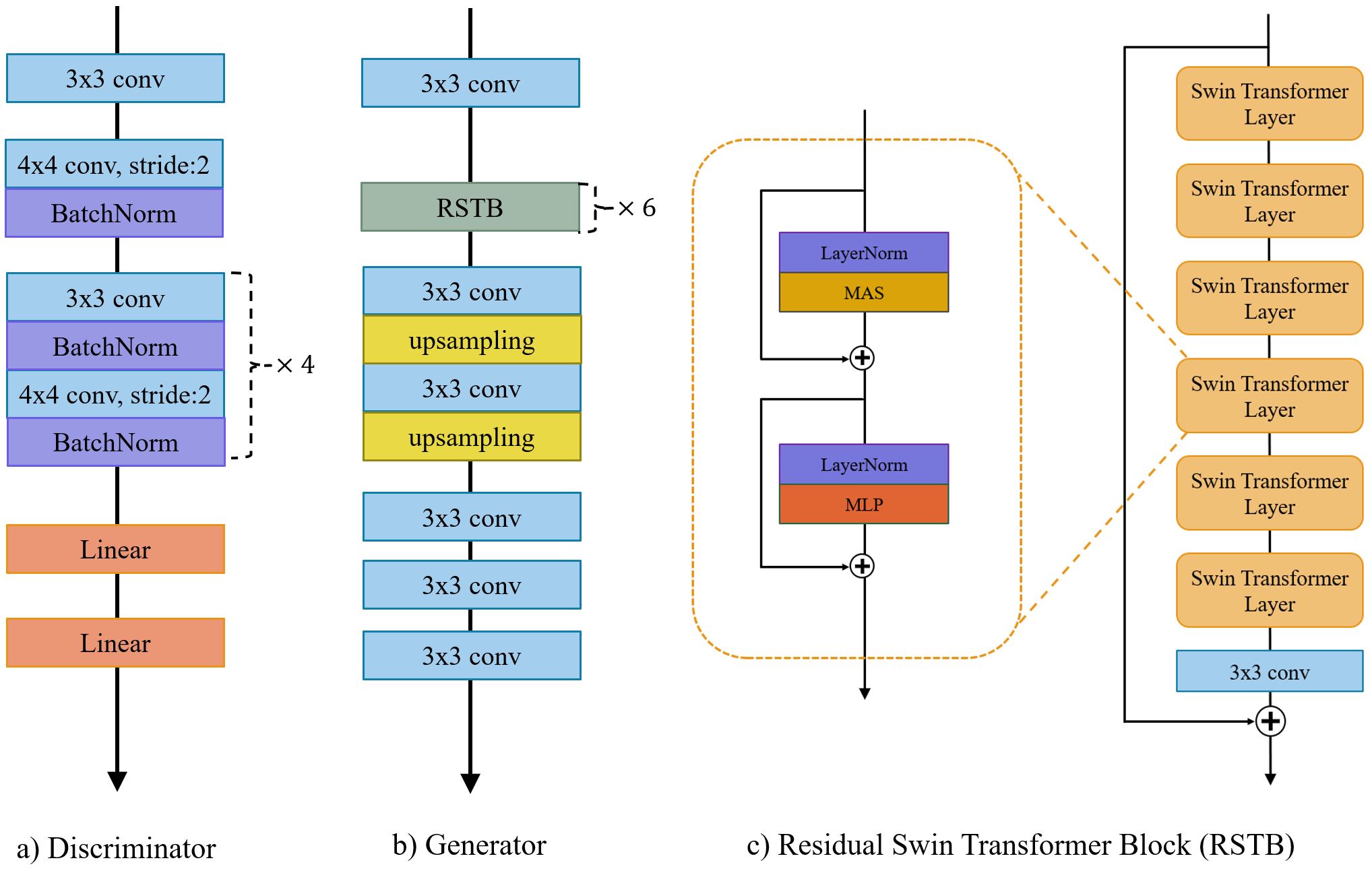}}
\caption{The detailed structure of the SwinIR. 
We omitted activation functions for simplicity.}
\label{fig:SwinIR_Structure}
\end{center}
\end{figure}

\clearpage

\section{Hyperparameters}\label{appendix:hyperparam}
\begin{table}[h]
\caption{Model hyperparameters in the first step.}
\begin{center}
\begin{tabular}{ccc}
\hline
\multicolumn{2}{c}{Hyperparameters}  &
\\ \hline
Basic setting & \begin{tabular}[c]{@{}c@{}}\# of iteration per scale\\ \# of intermediate layers\\ \# of channels\\ GT size\\ kernel size\\ stride\\ padding size \\ batch size\end{tabular}                                           & \begin{tabular}[c]{@{}c@{}}5000\\ 5\\ 64\\ 256\\ 3\\ 1\\ 1 \\ 2\end{tabular}                                \\ \hline
Optimizer     & \begin{tabular}[c]{@{}c@{}}type\\ betas\\ learning rate\\ gradient clip\\ weight of reconstruction loss \\ weight of KL loss \\ weight of discriminator loss \end{tabular} & \begin{tabular}[c]{@{}c@{}}Adam\\ (0.5, 0.999)\\ 0.0005\\ 5\\ 10\\ 1\\ 1\end{tabular} \\ \hline
Encoder & \begin{tabular}[c]{@{}c@{}} \# of blocks\end{tabular} & \begin{tabular}[c]{@{}c@{}} 2\end{tabular}  \\ \hline
Generator     & \begin{tabular}[c]{@{}c@{}}\# of Patch-VAE\\ \# of Patch-GAN\\ Scale factor\\ \end{tabular} & \begin{tabular}[c]{@{}c@{}}3\\ 6\\ 0.75\\  \end{tabular} \\ \hline
\end{tabular}
\end{center}
\end{table}

\begin{table}[h]
\caption{ESRGAN hyperparameters in the second and third steps.}
\begin{center}
\begin{tabular}{ccc}
\hline

\multicolumn{2}{c}{Hyperparameters}    & 2nd and 3rd step                                                                                                                                                \\ \hline
Basic setting & \begin{tabular}[c]{@{}c@{}}total iteration\\ \# of channels\\  GT size \\ batch size\end{tabular}                                     & \begin{tabular}[c]{@{}c@{}}100,000\\ 64\\ 128 \\ 16 \end{tabular} \\ \hline

Optimizer    & \begin{tabular}[c]{@{}c@{}} type \\ betas \\ learning rate\\ scheduling intervals \\ scheduling ratio \\ weight of perceptual loss\\  weight of reconstruction loss \\  weight of adversarial loss \end{tabular} & \begin{tabular}[c]{@{}c@{}} Adam \\ (0.9, 0.999)\\ 0.0001\\ 30k, 60k, 90k \\ 0.5\\ 1.0\\ 0.01\\ 0.005 \end{tabular} \\ \hline

Generator     & \begin{tabular}[c]{@{}c@{}}\# of RRDB \\ upscaling ratio\\ standard deviation of random noise \end{tabular} & \begin{tabular}[c]{@{}c@{}}23\\ 4\\ 0.1 \end{tabular} \\ \hline

\end{tabular}
\end{center}
\end{table}

\begin{table}[h]
\caption{SwinIR hyperparameters in the second and third steps.}
\begin{center}
\begin{tabular}{ccc}
\hline
\multicolumn{2}{c}{Hyperparameters}    & 2nd and 3rd step                                                                                                                                       \\ \hline

Basic setting & \begin{tabular}[c]{@{}c@{}}total iteration \\ \# of channels \\ GT size \\ batch size \end{tabular}       
              & \begin{tabular}[c]{@{}c@{}}130,000 \\ 180 \\ 96 \\ 20 \end{tabular}
\\ \hline

Optimizer    & \begin{tabular}[c]{@{}c@{}} type \\ betas \\ learning rate\\ scheduling intervals \\ scheduling ratio \\ weight of perceptual loss\\  weight of reconstruction loss \\  weight of adversarial loss \end{tabular} 
             & \begin{tabular}[c]{@{}c@{}} Adam \\ (0.9, 0.999)\\ 0.0001\\ 30k, 60k, 90k \\ 0.5\\ 1.0\\ 1.0\\ 0.1 \end{tabular}
\\ \hline

Generator     & \begin{tabular}[c]{@{}c@{}} \# of RSTB \\ \# of STL in each RSTB \\ \# of attention head \\ window size \\ standard deviation of random noise \end{tabular}
              & \begin{tabular}[c]{@{}c@{}} 6 \\ 6 \\  6 \\ 8 \\  0.1 \end{tabular}
\\ \hline

\end{tabular}
\end{center}
\end{table}

\end{document}